\documentclass{article} 
\usepackage{iclr2026_conference,times}

\usepackage{amsmath,amsfonts,bm}

\def\eqref#1{equation~\ref{#1}}

\def\1{\bm{1}}

\DeclareMathAlphabet{\mathsfit}{\encodingdefault}{\sfdefault}{m}{sl}
\SetMathAlphabet{\mathsfit}{bold}{\encodingdefault}{\sfdefault}{bx}{n}

\def\gN{{\mathcal{N}}}

\def\sR{{\mathbb{R}}}

\def\sZ{{\mathbb{Z}}}

\newcommand{\E}{\mathbb{E}}

\newcommand{\R}{\mathbb{R}}

\newcommand{\softmax}{\mathrm{softmax}}

\newcommand{\KL}{D_{\mathrm{KL}}}
\newcommand{\Var}{\mathrm{Var}}

\newcommand{\Cov}{\mathrm{Cov}}

\DeclareMathOperator*{\argmin}{arg\,min}

\DeclareMathOperator{\Tr}{Tr}

\renewcommand{\eqref}[1]{\textup{(\ref{#1})}}
\newcommand{\diag}{\mathrm{diag}}

\usepackage{hyperref}
\usepackage{url}

\renewcommand{\KL}[2]{D_{\mathrm{KL}}\left(#1 \,\Vert\, #2\right)}
\newcommand{\vect}[1]{\text{vec}(#1)}
\DeclareMathOperator{\chol}{chol}

\usepackage{amsthm}

\newtheorem{assumption}{Assumption}
\theoremstyle{definition}

\theoremstyle{remark}

\usepackage{algorithm}
\usepackage{algpseudocode}

\usepackage{booktabs}
\usepackage{subcaption}
\title{SoftWater: Class-aware rate allocation for softmax quantization}

\author{Joao V. Cavalcanti \& Ashia C. Wilson 
\\
Department of Electrical Engineering and Computer Science\\
Massachusetts Institute of Technology\\
Cambridge, MA 02139, USA \\
\texttt{\{caval,ashia\}@mit.edu}
}

\iclrfinalcopy 
\begin{document}

\maketitle

\begin{abstract}
Post-training quantization pipelines routinely leave the softmax output layer in high precision.
Yet in small LLMs with modern vocabularies, the head holds 15--30\% of all parameters, so a nominal ``2-bit'' model with an fp16 head can store several times as many bits per weight.
We pose softmax-layer quantization as a rate-distortion problem under the KL divergence between the original and quantized output distributions.
A second-order analysis reveals a class-aware geometry: quantization error is weighted jointly by feature covariance and class-specific softmax curvature.
A separability approximation replaces the $Kn\times Kn$ Cholesky with one $n\times n$ factorization rescaled per class, making the lattice encodable by successive interference cancellation, with both statistics from a single forward pass.
The resulting method, SoftWater, gives fine grids to frequent, low-variance classes and coarse grids to rare ones, a large gap under Zipfian token distributions.
Across five models from 1B to 32B, SoftWater outperforms the released WaterSIC quantizer (near-optimal under linear-layer WMSE but not output KL) at matched head rates on 59 of 60 test points, using none of that pipeline's refinements and cutting head-induced KL by $6.5\times$--$8.3\times$ at 2 bits.
On Llama-3.2-1B-Instruct with quantized bodies, a 2-bit head removes 45--60\% of stored bytes for a $2.9$--$3.7\%$ perplexity increase.
Because the class-side statistic comes from calibration data, matching calibration to the deployment domain gives the lowest KL on that domain throughout.
On a tied model, a 4-bit head is near-lossless and a 2-bit head costs under 4\% perplexity, making head quantization of such models practical.
\end{abstract}



\section{Introduction}

Post-training quantization (PTQ) replaces trained weights with low-precision ones while keeping the output close to the original.
For the linear layers that make up most of an LLM, the standard objective is a weighted mean-squared error (WMSE) on the layer output, $\E\Vert (W-\hat{W})X \Vert^{2}$, targeted by sequential rounding \cite{frantar_gptq_2023,frantar_optimal_2023,zhang_qronos_2026,chen_geometry_2026}, lattice codebooks \cite{chee_quip_2024,tseng_quip_2024,tseng_qtip_2025,savkin_nestquant_2025}, or learned clipping and rotations \cite{shao_omniquant_2024,liu_spinquant_2025,ashkboos_quarot_2024,shao_dartquant_2025}.
Most recently, WaterSIC \cite{lifar_watersic_2026} brought the WMSE problem within $0.255$ bits of its information-theoretic limit.

One layer does not fit this picture.
When a linear map feeds a softmax, the model outputs a probability distribution, so the distortion that matters is the KL divergence between the original and quantized outputs.
PTQ pipelines usually skip this layer, leaving the head in high precision and excluding it from the reported bits per weight.
That is costly: in small models with modern vocabularies the head holds $15$--$30\%$ of all parameters (Table~\ref{tab:head-share}), so a ``2-bit'' Llama-3.2-1B with an fp16 head stores nearly $5$ bits per weight.
Prior work on the output layer replaces it \cite{grave_efficient_2017,jean_using_2015,liu_csv-decode_2025,shao_vq-logits_2025,tranheden_flashhead_2026,lee_lfq_2026} rather than quantizing it under the KL metric.

\begin{table}[t]
\caption{Share of total parameters held by the output head (vocab $\times$ hidden) in modern small LLMs.}
\label{tab:head-share}
\begin{center}
\begin{tabular}{lrrrr}
\multicolumn{1}{c}{\bf Model} & \multicolumn{1}{c}{\bf Vocab} & \multicolumn{1}{c}{\bf Hidden} & \multicolumn{1}{c}{\bf Head (M)} & \multicolumn{1}{c}{\bf \% of total} \\
\hline
\rule{0pt}{2.6ex}Gemma-3-1B & 262{,}144 & 1152 & 302 & 30.2 \\
Qwen3-0.6B   & 151{,}936 & 1024 & 156 & 26.1 \\
Gemma-2-2B   & 256{,}128 & 2304 & 590 & 22.6 \\
Llama-3.2-1B & 128{,}256 & 2048 & 263 & 21.3 \\
Qwen3-1.7B   & 151{,}936 & 2048 & 311 & 18.1 \\
Gemma-3-4B   & 262{,}144 & 2560 & 671 & 15.6 \\
\hline
\end{tabular}
\end{center}
\end{table}

We treat the head as a rate-distortion problem under KL.
Where WMSE uses the error weight $I \otimes \Sigma_{X}$, a second-order expansion of the KL divergence gives $\E[(\diag(p)-pp^{\top}) \otimes XX^{\top}]$: the softmax Hessian ties every class to the input statistics.
A separable surrogate, the diagonal statistic $\bar{\lambda}_{k} = \E[p_{k}(1-p_{k})]$ paired with the usual feature covariance, gives class $k$ a grid scaled by $\bar{\lambda}_{k}^{-1/2}$ and column $i$ a grid scaled by $\vert \ell_{ii}\vert^{-1}$.
The resulting method, \emph{SoftWater}, gives fine grids to frequent, low-variance classes and coarse grids to rare ones; token distributions are Zipfian, so this gap is large.
Exploiting it costs one extra diagonal accumulator in the same calibration pass, and the decode format does not change.

The raw allocation is coarsest where the expansion is least trustworthy: classes with vanishing calibration probability.
We therefore smooth the calibration distribution with a prior, capping the grid spacing on those rows.
The smoothing weight interpolates between two regimes: at one end the calibration distribution sets the allocation, at the other every class gets equal weight and we recover WaterSIC.
SoftWater thus contains the WMSE treatment of the head as the case where the deployment distribution is assumed uniform.

Our contributions are:
\begin{itemize}
    \item \textbf{Problem formulation.} We pose softmax-layer quantization as a rate-distortion problem under output KL and derive the induced error metric. A dither argument and a separability assumption, which we validate empirically, reduce it to a class-rescaled lattice problem whose two statistics are read off one calibration pass (\S\ref{sec:problem-setup}--\ref{sec:softwater}).
    \item \textbf{The SoftWater algorithm.} The class-side scale is class frequency discounted by cross-context variability, so SoftWater spends bits on classes that are frequent and low-variance. A smoothing prior caps the spacing on classes the calibration set leaves uncovered, keeping the allocation inside the range where the Taylor expansion is valid (\S\ref{sub:ll-vs-softmax}--\ref{sub:accurate-proxy}).
    \item \textbf{Head quantization results.} Across five models from $1$B to $32$B parameters, SoftWater outperforms the released WaterSIC quantizer at matched head rates on $59$ of $60$ test points while using none of that pipeline's refinements, and transfers unchanged to models with released GuidedQuant bodies \cite{kim_guidedquant_2025}. The class-side factor carries domain information the covariance does not, letting a head be targeted at its deployment domain, and it cuts whole-model size by up to $60\%$ (\S\ref{sec:experiments}).
\end{itemize}

\section{Prior work}
\label{sec:prior-work}

We review the WMSE formulation of linear layer quantization, the WaterSIC algorithm whose SIC encoder we reuse, and recent work beyond MSE objectives.

\subsection{Linear layer quantization}
\label{sub:ll-quantization}

A linear layer takes an activation $X\in\R^{n}$ and a weight matrix $W\in\R^{K\times n}$ and outputs $z=WX$.
Quantization looks for $\hat{W}\in\R^{K\times n}$ minimizing the distortion
\begin{align}
    D
    &= \frac{1}{K n}\E\Vert (\hat{W}-W)X \Vert_{F}^{2}
    = \frac{1}{K n}\sum_{k=1}^{K}(\hat{W}_{k}-W_{k})\Sigma_{X}(\hat{W}_{k}-W_{k})^{\top},
    \label{eq:linear-layer-distortion}
\end{align}
where $W_{k}$ and $\hat{W}_{k}$ are rows and $\Sigma_{X}=\E_{X}[XX^{\top}]$.
Most calibration-based PTQ methods target \eqref{eq:linear-layer-distortion}: GPTQ and its successors by sequential rounding \citep{frantar_gptq_2023,frantar_optimal_2023,zhang_qronos_2026,chen_geometry_2026}, QuIP-style methods by incoherence processing and structured codebooks \citep{chee_quip_2024,tseng_quip_2024,tseng_qtip_2025,savkin_nestquant_2025}.

\subsection{The WaterSIC algorithm}
\label{sub:watersic}

\citet{lifar_watersic_2026} solve \eqref{eq:linear-layer-distortion} within $0.255$ bits of the information-theoretic limit \citep[Theorem 3.3]{lifar_watersic_2026} by allocating rates \emph{per column} from $\Sigma_{X}$.

Let $L$ be the lower triangular Cholesky factor of $\Sigma_{X}=LL^{\top}$.
WaterSIC restricts $\hat{W}_{k}$ to a scaled integer lattice $\sZ^{1\times n}A$, where $A=\diag(\alpha_{1},\cdots,\alpha_{n})$ with $\vert A \vert^{1/n}=\bar{\alpha}$: the lattice has the same point density $\bar{\alpha}^{-n}$ as $\bar{\alpha}\sZ^{1\times n}$, but each $\alpha_{i}$ sets the grid resolution of column $i$, which entropy coding turns into a per-column bit rate.
The objective reads
\begin{align}
    D
    = \frac{1}{K n}\sum_{k=1}^{K}\Vert (\hat{W}_{k}A-W_{k})L \Vert^{2},
    \label{obj:wmse-watersic}
\end{align}
with $\hat{W}_{k}$ restricted to $\sZ^{1\times n}$, an NP-hard problem.
WaterSIC solves it approximately with successive interference cancellation (SIC), which exploits the triangular structure of $L$: columns are encoded in sequence by scalar equations that account for previously quantized entries.
All rows are processed identically, since \eqref{obj:wmse-watersic} decouples them.

It remains to set the scales $A$.
Assume the rows $W_{k}$ are iid $\gN(0,\sigma_{W}^{2}I)$ and $\sigma_{W}/\bar{\alpha} \to \infty$.
Then the quantization error approaches a uniform distribution \citep[Lemma 3.2, Appendix B]{lifar_watersic_2026}, and with per-column entropy coding the rate approaches, as $K$ grows,
\begin{align}
    R_{\text{SIC}}
    \approx 
    R_{\text{High-Rate}}(D,\Sigma_{X})
    +\frac{1}{2}\log\left( \frac{2\pi e}{12} \right)
    +\frac{1}{2}\log\left( \frac{\frac{1}{n}\sum_{i=1}^{n}(\alpha_{i}\ell_{ii})^{2}}{\prod_{i=1}^{n}(\alpha_{i}\ell_{ii})^{2/n}} \right),
    \label{approx:SIC-rate}
\end{align}
where $R_{\text{High-Rate}}(D,\Sigma_{X})$ is the optimal rate and $\ell_{ii}$ is the $i$-th diagonal entry of $L$.
The choice
\begin{align}
    \alpha_{i}
    = \frac{c_{\text{WS}}}{\vert \ell_{ii} \vert},
    \quad i = 1,\cdots, n,
    \label{eq:alpha-watersic}
\end{align}
kills the third term and attains the $(1/2)\log(2\pi e/12)\approx 0.255$ bit gap; here $c_{\text{WS}}>0$ sets the lattice density, and $c_{\text{WS}}=\bar{\alpha} \vert L\vert^{1/n}$ recovers the density $\bar{\alpha}^{-n}$.

The released pipeline adds several modifications \cite[see \S 4]{lifar_watersic_2026}.
We keep the two that any unequal-rate scheme needs: \emph{entropy coding}, which turns the integer codes into a bitstream and is what makes unequal per-column rates realizable, and \emph{rate assignment}, which tunes $c_{\text{WS}}$ by the secant method since the post-coding rate cannot be set directly.
The rest are quality refinements orthogonal to rate allocation, and we do not use them, so the gains in \S\ref{sec:experiments} are attributable to the allocation alone and are in principle additive with them.
One deserves mention: dead feature erasure targets near-zero-variance coordinates in early-layer inputs, and the released quantizer applies it to the head too; because the head reads a well-conditioned final hidden state, we use nominal damping instead ($\delta = 10^{-6}\,\mathrm{mean}(\diag \Sigma_{X})$).
Activation drift and residual stream correction are vacuous for a terminal layer that runs on a fixed body and writes into nothing, so we omit them.

\subsection{Quantization metrics beyond MSE}
\label{sub:beyond-mse}

WMSE is the standard metric in model quantization \citep[e.g.,][]{nagel_up_2020,frantar_gptq_2023}, mainly because it is tractable: it decouples rows, admits tools like Cholesky decompositions, connects directly to rate-distortion theory, and needs forward passes only.
But it is a proxy for output behavior, and recent work has looked for better metrics; \citet{lifar_watersic_2026} list approximating a target perplexity or KL loss as their first open direction.
\citet{lee_lfq_2026} swap MSE for the cross entropy between full-precision and quantized logits in the final block before the head.
\citet{kim_guidedquant_2025} weight the layerwise objective by end-loss guidance, made tractable by assuming Fisher coefficients coupling different output channels vanish, so the objective again decouples across rows; this is the end-to-end analogue of the cross-row decorrelation we derive in \S\ref{sub:tractable-proxy}, which for the softmax layer we can justify rather than assume, and their released pipeline produces the quantized bodies of \S\ref{sec:experiments}.
\citet{tseng_model-preserving_2026} build YAQA on tractable Hessian sketches of end-to-end KL, also decomposing the Hessian as a Kronecker product, but target adaptive rounding for body layers, whereas we derive the layer-local KL geometry of the softmax head and its rate allocation.
A separate line replaces the head itself \citep{grave_efficient_2017,jean_using_2015,liu_csv-decode_2025,shao_vq-logits_2025,tranheden_flashhead_2026}; we keep the architecture and quantize the existing head.

\section{Problem setup}
\label{sec:problem-setup}

We consider a softmax layer that takes an input $X \in \R^{n}$ with $n$ features and produces a distribution $p = \softmax(z) \in \R^{K}$ over $K$ classes, where $z = WX$ are the logits, $W \in \R^{K\times n}$ are the weights with rows $W_{k}\in \R^{1\times n}$, and
\begin{align*}
    p_{k}
    = \softmax(z)_{k}
    = \frac{\exp(z_{k})}{\sum_{l=1}^{K}\exp(z_{l})}.
\end{align*}
We treat $X$ as a random vector drawn from the distribution of layer inputs induced by the data the model is deployed on, and estimate expectations with respect to it from a calibration sample.
Both $z$ and $p$ are therefore functions of $X$; we suppress the dependence except where it matters.

For LLM heads, $K$ is the vocabulary size and $W$ can hold a large fraction of all parameters, and is often tied to the input embedding (Table~\ref{tab:head-share}).
To mitigate this cost, we replace $W$ by a matrix $\hat{W}$ whose entries are described by a budget of $R$ bits per weight.
Let $\Delta = \hat{W}-W$ and $\delta = \Delta X$ denote the weight and logit perturbations, with rows $\Delta_{k} \in \R^{1\times n}$ and entries $\delta_{k} = \Delta_{k} X$.
The class distribution of the quantized model is then $q \in \R^{K}$, defined by $q_{k} = \softmax(z+\delta)_{k}$.
Our goal is to find, for a given rate $R$, the $\hat{W}$ that minimizes the discrepancy between $p$ and $q$, which \S\ref{sub:objective-derivation} formalizes as an expected KL divergence.

\section{SoftWater: per-Column and per-class rate allocation}
\label{sec:softwater}

The KL geometry of a softmax layer induces rate allocation along two axes: input features and output classes. 
SoftWater allocates along both axes. 
A separability approximation makes the resulting lattice encodable by SIC at $O(n^{3})$ cost.
We derive its objective, reduce it to a tractable problem, add a smoothing step that keeps the proxy accurate, and contrast linear layer and softmax quantization.

\subsection{Linear layer vs softmax quantization}
\label{sub:ll-vs-softmax}

In \S\ref{sub:watersic} we saw that SIC's error geometry over the Cholesky factor $L$, combined with entropy coding, points to grid spacing proportional to $\vert \ell_{ii} \vert^{-1}$.
Our goal in this section is to show that, for softmax quantization, the grid spacing should instead be proportional to $(\bar{\lambda}_{k}^{1/2} \vert \ell_{ii} \vert)^{-1}$, where
\begin{align*}
    \bar{\lambda}_{k}
    = \E_{X}[p_{k}(1-p_{k})]
    = \E_{X}[p_{k}](1-\E_{X}[p_{k}]) - \Var[p_{k}].
\end{align*}
That is, the grid spacing is inversely proportional to the square root of class frequency discounted by cross-context variance, the spread of $p_{k}$ across contexts.
Therefore, through $(\bar{\lambda}_{k}^{1/2} \vert \ell_{ii} \vert)^{-1}$,
\begin{center}
    \emph{SoftWater corrects for in-feature geometry, class frequency and class variance}.
\end{center}
In particular, SoftWater allocates
\begin{itemize}
    \item \textbf{more bits to frequent classes than infrequent classes};
    \item \textbf{more bits to low-variance classes than high-variance classes}.
\end{itemize}

The size of the gap between the two grid spacing choices belongs to the class distribution, not the algorithm.
For a near-uniform distribution the $\bar{\lambda}_{k}$ are nearly equal and SoftWater recovers WaterSIC.
For the heavy tails of natural-language vocabularies, $\E_{X}[p_{k}]$ spans many orders of magnitude and the spread in $\bar{\lambda}_{k}^{-1/2}$ is large, bounded only by the smoothing floor of \S\ref{sub:accurate-proxy}.
This is why the class side, not the feature side, is where a softmax layer's rate budget is won or lost.
Solving the softmax problem through WMSE therefore amounts to assuming identically distributed classes.

\subsection{Objective derivation}
\label{sub:objective-derivation}

Linear layer quantization measures distortion by the output WMSE \citep{frantar_gptq_2023,lifar_watersic_2026}.
A softmax output has more structure: it lies on the probability simplex and is a distribution.
We therefore use the KL divergence, which captures changes in that distribution directly, and look for $\hat{W}$ minimizing
\begin{align*}
    \KL{p}{q}
    &= \sum_{k}p_{k}(\log p_{k} -\log q_{k}).
\end{align*}
Writing
\begin{align*}
    \log p_{k}
    = z_{k} - \log \sum_{l=1}^{K}\exp(z_{l}),
    \quad
    \log q_{k}
    = z_{k}+\delta_{k} - \log \sum_{l=1}^{K}\exp(z_{l}+\delta_{l}),
\end{align*}
and substituting gives
\begin{align*}
    \KL{p}{q}
    &= \log \sum_{l=1}^{K}\exp(z_{l}+\delta_{l})
    -\log \sum_{l=1}^{K}\exp(z_{l})
    -p^{\top}\delta.
\end{align*}
Let $f(\delta) = \KL{p}{q}$, with
\begin{align*}
    \nabla f(\delta) 
    &= \softmax(z+\delta) -p \quad \text{and} \quad 
    \nabla^{2} f(\delta)
    = \diag(q) -qq^{\top}.
\end{align*}
Since $f(0)=0$ and $\nabla f(0)=0$, a second-order Taylor expansion at $\delta = 0$ gives
\begin{align}
    \KL{p}{q}
    &= f(0) + \nabla f(0)^{\top}\delta + \frac{1}{2}\delta^{\top}\nabla^{2}f(0)\delta + O(\Vert \delta \Vert^{3})
    \nonumber\\
    &\approx \frac{1}{2}X^{\top}\Delta^{\top} \nabla^{2}f(0) \Delta X
    \nonumber\\
    &= \frac{1}{2} \vect{\Delta}^{\top}(\nabla^{2}f(0) \otimes XX^{\top})\vect{\Delta},
    \label{approx:kl-taylor}
\end{align}
where $\vect{\Delta}\in \R^{K n}$ stacks the rows of $\Delta$ and  
\begin{align}
    \nabla^{2}f(0) 
    = \diag(p) -pp^{\top}.
    \label{eq:hessian-f}
\end{align}
The expansion \eqref{approx:kl-taylor} holds as long as the logit perturbation stays small; \S\ref{sub:accurate-proxy} enforces this and \S\ref{sub:separability-exp} verifies the reduction that follows.
The objective is then
\begin{align}
    \E_{X} \left[
    \frac{1}{2} \vect{\Delta}^{\top}(\nabla^{2}f(0) \otimes XX^{\top})\vect{\Delta}
    \right]
    =
    \frac{1}{2} \vect{\Delta}^{\top}\E_{X}[\nabla^{2}f(0) \otimes XX^{\top}]\vect{\Delta}.
    \label{obj:expected-kl-taylor}
\end{align}
The error weight is thus $\E_{X}[\nabla^{2}f(0) \otimes XX^{\top}]$, a $Kn\times Kn$ matrix coupling classes and features.
Encoding against this directly would need a Cholesky factorization of that $Kn \times Kn$ matrix, costing $O(K^{3}n^{3})$ time and $O(K^{2}n^{2})$ memory.
We need a tractable approximation, which its structure allows.

\subsection{A tractable proxy}
\label{sub:tractable-proxy}

The obstacle is that the expectation couples $\nabla^{2}f(0)$ and $XX^{\top}$, giving sample-dependent off-diagonal terms and a full matrix.
If we can decouple them, then $\chol(A \otimes B)=\chol(A) \otimes \chol(B)$ splits the factorization into two smaller ones.

Under SIC, rows are quantized independently against the same $L$, with sequential correction along coordinates within each row: errors are coupled \emph{within} a row, not \emph{across} rows.
Per-entry subtractive dither makes this precise.
Under the dither ensemble the errors are zero-mean and uncorrelated across rows, and the dither is drawn independently of the data, so the expectations over $X$ and over $\Delta$ factor.
The cross-terms $-p_{k}p_{l}$, $k \neq l$, then contribute nothing, and the class-side diagonal is exact:
\begin{align}
    \E_{X,\Delta}[\vect{\Delta}^{\top}(\nabla^{2}f(0) \otimes XX^{\top})\vect{\Delta}]
    = \E_{X,\Delta}[\vect{\Delta}^{\top}(\lambda \otimes XX^{\top})\vect{\Delta}],
    \label{eq:diagonal-dither}
\end{align}
where $\lambda$ is the softmax curvature
\begin{align}
    \lambda
    = \diag(p\odot(1-p)).
    \label{def:softmax-curvature}
\end{align}
Replacing the error weight in \eqref{obj:expected-kl-taylor} by its class-side diagonal gives
\begin{align}
    \frac{1}{2} \vect{\Delta}^{\top}\E_{X}[\lambda \otimes XX^{\top}]\vect{\Delta},
    \label{obj:dither}
\end{align}
again a deterministic function of the perturbation.
Dithering is thus the device that identifies \eqref{obj:dither} as the right target; empirically, undithered errors already behave as if uncorrelated across rows, so we drop it from the final scheme and leave it as an optional tweak.

The matrix $\E_{X}[\lambda\otimes XX^{\top}]$ is block diagonal, so $K$ Cholesky decompositions of size $n$ replace one of size $K n$.
Except for very small $K$ that is still too expensive, so we assume that $\lambda$ and $XX^{\top}$ are \emph{separable}.

\begin{assumption}[Separability]
\label{ass:separability}
    The softmax curvature $\lambda$ and $XX^{\top}$ are uncorrelated:
    \begin{align}
        \E_{X}[\lambda \otimes XX^{\top}]
        = \E_{X}[\lambda] \otimes \E_{X}[XX^{\top}].
        \label{approx:separability}
    \end{align}
\end{assumption}

Assumption \ref{ass:separability} holds only approximately; we bound the error below and test it in \S\ref{sub:separability-exp}.
It lets the two Kronecker factors be estimated independently: the covariance $\Sigma_{X}$ separately from the expected curvature
\begin{align}
    \bar{\lambda}
    = \E_{X}[\lambda]
    = \E_{X}[\diag(p\odot(1-p))].
    \label{def:softmax-curvature-expected}
\end{align}
Moreover, $\bar{\lambda}$ is just $K$ scalars, accumulated from the output distributions the calibration forward pass already produces: SoftWater needs no backward pass and no second pass over the data, and costs one $K$-vector alongside the $n \times n$ covariance, both from the same forward pass.
The Cholesky factor $L$ of $\Sigma_{X}$ is computed once and rescaled $K$ times to give the factor $\mathcal{L}=\bar{\lambda}^{1/2} \otimes L$ of $\bar{\lambda}\otimes\Sigma_{X}$.
The distortion metric is then
\begin{align}
    D =
    \frac{1}{2}\vect{\Delta}^{\top}\mathcal{L}\mathcal{L}^{\top}\vect{\Delta},
    \label{obj:softwater}
\end{align}
a WMSE over the joint index $(k,i)$, which SIC minimizes directly on $\mathcal{L}$.

The induced grid spacing for class $k$ and column $i$ is
\begin{align}
    \alpha_{(k-1)n + i}
    = \frac{c_{\text{SW}}}{\bar{\lambda}_{k}^{1/2}\vert \ell_{ii} \vert},
    \quad k = 1,\cdots, K,
    \quad i = 1,\cdots, n,
    \label{eq:alpha-softwater}
\end{align}
where $c_{\text{SW}}>0$ again sets the point density.
Letting
\begin{align}
    \mathcal{B}
    &= 
    \diag(\bar{\lambda}_{1}^{-1/2}, \cdots, \bar{\lambda}_{K}^{-1/2}),
    \label{eq:sw-classwide-rescalers}
    \\
    \mathcal{A}
    &= 
    \diag(\vert \ell_{11} \vert^{-1}, \cdots, \vert \ell_{nn} \vert^{-1}),
    \label{eq:sw-activationwise-rescalers}
\end{align}
and choosing
\begin{align}
    c_{\text{SW}}
    = \bar{\alpha} \vert \mathcal{B} \vert^{-1/K} \vert \mathcal{A} \vert^{-1/n},
    \label{eq:sw-csw}
\end{align}
gives point density $\bar{\alpha}^{-Kn}$.
SoftWater therefore uses $c_{\text{SW}}\mathcal{B}\sZ^{K\times n}\mathcal{A}$, rescaled on the class side as well as the feature side.
The hyperparameter $c_{\text{SW}}$ is the sole rate knob: it scales the lattice uniformly, raises the post-coding rate monotonically, and is set by secant search (\S\ref{sub:watersic}).

\paragraph{Remark.} 
We notice that $\mathcal{A}$ and $\mathcal{B}$ are not just column and row rescalers; they change the grid spacing itself, redistributing rate across classes through the coupling with $c_{\text{SW}}$.
In contrast, plain column and row rescalers applied after coding change reconstruction but not rate.

\paragraph{On the separability assumption.}
Although $\lambda_{k}=\lambda_{k}(X)$ depends on $X$, separability does not require independence between class probabilities and the input.
It requires only that the class-dependent curvature be weakly correlated with the second-order geometry of $X$.
The error in the $k$-th block is
\begin{align}
\E_{X}[\lambda_{k}(X)XX^\top]-\bar{\lambda}_{k}\Sigma_{X}
=
\E_{X} \left[
(\lambda_{k}(X)-\bar{\lambda}_{k})(XX^\top-\Sigma_{X})
\right],
\label{eq:separability-error}
\end{align}
so the approximation is accurate when examples with greater curvature on class $k$ do not have systematically different feature covariance.
More directly, for the $k$-th row $\Delta_{k}$ of the perturbation,
\begin{align}
\Delta_{k}
\left(
\E_{X}[\lambda_{k}(X)XX^\top]-\bar{\lambda}_{k}\Sigma_{X}
\right)
\Delta_{k}^{\top}
=
\Cov_{X}\left(
\lambda_{k}(X),(\Delta_{k} X)^2
\right).
\label{eq:directional-separability-error}
\end{align}
The proxy therefore need not match each block in every direction.
It is enough that class uncertainty be weakly correlated with feature energy along the error directions the algorithm produces.
This is plausible when representations are normalized and their geometry is stable across contexts.
If $\Vert X \Vert^2$ is constant, for instance, the exact and separable blocks have the same trace,
\begin{align*}
\Tr\left(\E_{X}[\lambda_{k}(X)XX^\top]\right)
= \bar{\lambda}_{k}\Tr(\Sigma_{X}),
\end{align*}
so separability preserves the total curvature per class and drops only class-dependent anisotropy.

Such factorizations are common for tractable Kronecker-factored curvature.
\citet{martens_optimizing_2015} reach the same approximation from natural gradient descent, where the Fisher matrix defines the KL trust region and its output-layer block is the Hessian of our objective; they need separability to invert the preconditioner via $(A\otimes B)^{-1} = A^{-1}\otimes B^{-1}$.
\citet{tseng_model-preserving_2026} reach Kronecker approximations in weight-only PTQ, but their treatment is agnostic to the output layer and does not exploit softmax structure, so they approximate the end-to-end error Hessian and obtain their factors by power iteration, whereas ours come from an explicit analytical expression.
In \S\ref{sub:separability-exp} we measure how far our factors sit from the Frobenius-optimal Kronecker factorization of the same error weight, and find the proxy underestimates the true distortion by at most 10\% across models, rates, and datasets.

\subsection{A tractable and accurate proxy: smoothing}
\label{sub:accurate-proxy}

The proxy \eqref{obj:softwater} is accurate only if the above assumptions hold, which we verify in \S\ref{sec:experiments}, and if the Taylor expansion of $\KL{p}{q}$ stays accurate, which requires a small logit perturbation $\delta$.

Grid spacing under our metric is proportional to $(\bar{\lambda}_{k}^{1/2} \vert\ell_{ii}\vert)^{-1}$, so rare classes get coarser grids.
If a class is absent from the calibration data, then $p_{k}\ll 1$ and $\bar{\lambda}_{k}^{1/2} \ll 1$, the grid is far too coarse, the weights are badly perturbed, and \eqref{approx:kl-taylor} breaks down.
We therefore smooth $p$ to lower bound $p_{k}$, computing
\begin{align}
    \tilde{\lambda}
    = \E_{X}[\diag(\tilde{p}\odot (1-\tilde{p}))],
    \label{def:softmax-curvature-smoothed}
\end{align}
where, given $\epsilon\in(0,1)$,
\begin{align}
    \tilde{p}
    = (1-\epsilon)p + \epsilon/K
    \label{eq:uniform-smoothing}
\end{align}
mixes $p$ with a uniform distribution over the $K$ classes.
Since $\tilde{p}_{k} \geq \epsilon/K$ pointwise, $\tilde{\lambda}_{k} \geq \tfrac{\epsilon}{K}(1-\tfrac{\epsilon}{K})$, while $\tilde{\lambda}_{k} \leq 1/4$ always.
The ratio between the coarsest and finest class-side spacings is therefore at most $\tfrac{1}{2}\sqrt{K/\epsilon}$ up to an $O(\epsilon/K)$ correction, so $\epsilon$ caps the class-side rate spread at $\log_{2}$ of that quantity, whatever the calibration data.
Uncovered classes thus keep logit perturbations inside the regime where \eqref{approx:kl-taylor} is valid.

So $\epsilon$ interpolates the curvature.
At $\epsilon=0$, $\tilde{\lambda}$ comes from the calibration data alone \eqref{def:softmax-curvature-expected}.
At $\epsilon=1$, it weighs every class equally and recovers WaterSIC.
We fix $\epsilon=0.1$ as default, and $\tilde{\lambda}$ replaces $\bar{\lambda}$ in the grid spacings \eqref{eq:alpha-softwater}, the rescalers $\mathcal{B}$, and the constant $c_{\text{SW}}$.

\begin{algorithm}
\caption{SoftWater weight-only quantization of softmax}
\label{alg:softwater}
\begin{algorithmic}[1]
\Require $W \in \mathbb{R}^{K \times n}$, $\Sigma_{X}\succeq 0$, $\tilde{\lambda}=\E_{X}[\diag(\tilde{p}\odot (1-\tilde{p}))]$, $\bar{\alpha} > 0$.
\Ensure $\alpha \in \mathbb{R}_{+}^{n}$, $\beta \in \mathbb{R}_{+}^{K}$, $B_{1},\dots,B_{n} \in \{0,1\}^{*}$ s.t.\ $\hat{W} = \operatorname{diag}(\beta)\, Z_{\mathrm{SIC}}\operatorname{diag}(\alpha)$.
\State Compute lower-triangular $L \in \mathbb{R}^{n\times n}$ s.t. $\Sigma_{X} = LL^\top$ 
\Comment{Cholesky}
\State $\beta_{k} \gets \frac{(\prod_{l}\tilde{\lambda}_{l}^{1/2})^{1/K}}{\tilde{\lambda}_{k}^{1/2}}, \quad \forall k \in [K]$ 
\Comment{class-side scales}
\State $\alpha_{i} \gets \bar{\alpha} \cdot \frac{|L|^{1/n}}{|\ell_{ii}|}, \quad \forall i \in [n]$ 
\Comment{feature-side water filling; $\ell_{ii}$ is the $i$th diagonal element of $L$}
\State $R \gets WL$
\For{$i = n : 1$}
    \State $Z_{\mathrm{SIC},:,i} \gets \operatorname{round}\!\big( R_{:,i} \oslash (\alpha_{i} \ell_{ii}\, \beta) \big)$ \Comment{$\oslash$: elementwise division by the $K$-vector $\alpha_{i} \ell_{ii}\, \beta$}
    \State $R \gets R - \alpha_{i} \big(\beta \odot Z_{\mathrm{SIC},:,i}\big) \cdot L_{i,:}$ \Comment{$\odot$: elementwise product; $L_{i,:}$ is the $i$th row of $L$}
\EndFor
\For{$i = 1 : n$}
    \State $B_{i} \gets \mathrm{EC}\big(Z_{\mathrm{SIC},:,i}\big)$ \Comment{entropy coding of the $i$th column}
\EndFor
\end{algorithmic}
\end{algorithm}

\section{Experiments}
\label{sec:experiments}

We evaluate SoftWater on five models from $1$B to $32$B parameters across the Llama and Qwen families: the head alone against an fp16 body (\S\ref{sub:head-only}), then fully quantized models with released GuidedQuant bodies (\S\ref{sub:quantized-body}), evaluated downstream in \S\ref{sub:zeroshot}.
Three further experiments probe the method: \S\ref{sub:domain-targeted} targets calibration at a deployment domain, \S\ref{sub:class-side-statistics} asks what the class-side statistic needs to be, and \S\ref{sub:separability-exp} tests separability.
Our arm runs the plain allocation with no post-hoc rescaler optimization, LMMSE shrinkage, or finetuning; the baseline is the full released WaterSIC pipeline minus finetuning.

\subsection{Head-only isolation}
\label{sub:head-only}

We quantize the heads of five models to 2, 3 and 4 bits and keep their bodies in fp16; Llama-3.2-1B and Qwen3-1.7B tie head and embedding, so a separate fp16 embedding is kept.
The baseline is the official WaterSIC release\footnote{Code downloaded from \url{https://github.com/egorlifar/watersic}.} without finetuning, so both arms are purely layerwise, with nominal damping $\delta = 10^{-6}\,\mathrm{mean}(\diag \Sigma_{X})$.
Both are calibrated on 128 sequences of 1024 tokens from WikiText2 (WT2), and we measure perplexity (PPL) and $\KL{p_{\text{base}}}{p_{\text{quant}}}$ on WT2 and C4.

\begin{table}
\caption{The body is kept in fp16 and only the head matrix is quantized.
KL is measured against the all-fp16 model on identical token streams.
WS: WaterSIC w/o FT.
Parentheses after each model name give its vocabulary size; on PPL cells, the increase over the fp16 row of the same model; on SW KL cells, the WS/SW KL ratio at that rate.
Best value per head group in bold.}
\label{tab:head-only-fp16-body}
\begin{center}
\scriptsize
\renewcommand{\arraystretch}{0.9}
\setlength{\tabcolsep}{3pt}
\begin{tabular}{lrllll}
\multicolumn{1}{c}{\bf Method} & \multicolumn{1}{c}{\bf Head} & \multicolumn{1}{c}{\bf WT2 PPL ($\downarrow$)} & \multicolumn{1}{c}{\bf WT2 KL ($\downarrow$)} & \multicolumn{1}{c}{\bf C4 PPL ($\downarrow$)} & \multicolumn{1}{c}{\bf C4 KL ($\downarrow$)}
\\ \hline
\multicolumn{6}{c}{\rule{0pt}{2.6ex}\bf Llama-3.2-1B (128{,}256)} \\ \hline
\rule{0pt}{2.6ex}fp16 & 16 & 11.313 & 0 & 15.614 & 0 \\
\cline{1-6}
\rule{0pt}{2.6ex}WS & 2 & 12.049 ($+$6.5\%) & 0.06192 & 17.283 ($+$10.7\%) & 0.09957 \\
\textbf{SW (ours)} & 2 & \textbf{11.419 ($+$0.9\%)} & \textbf{0.00957 (6.5$\times$)} & \textbf{16.035 ($+$2.7\%)} & \textbf{0.02834 (3.5$\times$)} \\
\cline{1-6}
\rule{0pt}{2.6ex}WS & 3 & 11.469 ($+$1.4\%) & 0.01364 & 15.965 ($+$2.2\%) & 0.02176 \\
\textbf{SW (ours)} & 3 & \textbf{11.342 ($+$0.3\%)} & \textbf{0.00206 (6.6$\times$)} & \textbf{15.704 ($+$0.6\%)} & \textbf{0.00572 (3.8$\times$)} \\
\cline{1-6}
\rule{0pt}{2.6ex}WS & 4 & 11.356 ($+$0.4\%) & 0.00319 & 15.713 ($+$0.6\%) & 0.00526 \\
\textbf{SW (ours)} & 4 & \textbf{11.320 ($+$0.1\%)} & \textbf{0.00047 (6.8$\times$)} & \textbf{15.640 ($+$0.2\%)} & \textbf{0.00138 (3.8$\times$)} \\
\hline
\multicolumn{6}{c}{\rule{0pt}{3.2ex}\bf Qwen3-32B (151{,}936)} \\ \hline
\rule{0pt}{2.6ex}fp16 & 16 & 8.452 & 0 & 12.269 & 0 \\
\cline{1-6}
\rule{0pt}{2.6ex}WS & 2 & 8.756 ($+$3.6\%) & 0.05835 & 13.029 ($+$6.2\%) & 0.08106 \\
\textbf{SW (ours)} & 2 & \textbf{8.498 ($+$0.5\%)} & \textbf{0.00703 (8.3$\times$)} & \textbf{12.479 ($+$1.7\%)} & \textbf{0.01596 (5.1$\times$)} \\
\cline{1-6}
\rule{0pt}{2.6ex}WS & 3 & 8.500 ($+$0.6\%) & 0.01246 & 12.439 ($+$1.4\%) & 0.01683 \\
\textbf{SW (ours)} & 3 & \textbf{8.464 ($+$0.1\%)} & \textbf{0.00147 (8.5$\times$)} & \textbf{12.315 ($+$0.4\%)} & \textbf{0.00341 (4.9$\times$)} \\
\cline{1-6}
\rule{0pt}{2.6ex}WS & 4 & 8.477 ($+$0.3\%) & 0.00295 & 12.302 ($+$0.3\%) & 0.00397 \\
\textbf{SW (ours)} & 4 & \textbf{8.457 ($+$0.1\%)} & \textbf{0.00035 (8.4$\times$)} & \textbf{12.275 ($+$0.1\%)} & \textbf{0.00081 (4.9$\times$)} \\
\hline
\end{tabular}
\end{center}
\end{table}

Table~\ref{tab:head-only-fp16-body} shows two of the five models; Table~\ref{tab:head-only-fp16-body-full} gives the rest.
Every arm lands within $0.005$ bits of its target rate, and SoftWater wins $59$ of $60$ cells; in all five models a 2-bit SoftWater head outperforms a 3-bit WaterSIC head on WT2 KL, so the method buys roughly one bit.
The KL advantage is $6.5\times$ to $8.3\times$ on WT2 at 2 bits and nearly flat in head rate, but smaller out of domain\footnote{In-domain means the distribution of the calibration data. Out-of-domain is analogous.} ($3.5\times$--$5.1\times$ on C4); perplexity behaves differently, with the gap to fp16 nearly closing at 4 bits.

\subsection{Quantized body, tied head}
\label{sub:quantized-body}

We next evaluate fully quantized Llama-3.2-1B-Instruct, whose tied head means the quantized weights are also the embedding, so head quantization hits both ends of the model at once.
Since the WaterSIC repository has no quantized bodies, we use released checkpoints\footnote{\url{https://huggingface.co/jusjinuk/Llama-3.2-1B-Instruct-2bit-GuidedQuant-QTIP}.} (BlockLDLQ under GuidedQuant Hessians~\cite{kim_guidedquant_2025} in the QTIP trellis format~\cite{tseng_qtip_2025}), decoded to dense fp16 with a standalone reimplementation verified against the reference forward.
Both quantizers calibrate on hidden states from the frozen quantized-body model, so $\Sigma_{X}$ and $\tilde{\lambda}$ reflect what the head sees at deployment, and neither activation drift nor residual compensation applies (\S\ref{sub:watersic}).

\begin{table}
\caption{Perplexity (PPL) on WikiText2 (WT2) and C4, and KL divergence from baseline: fp16-head, body quantized at 2 and 4 bits with GuidedQuant.
Body and head columns give respective rates.
WS: WaterSIC w/o FT.
Parentheses give increase over the fp16-head row of the same body.
Model size and reduction ($\Delta$size) after quantizing the tied head.
Best value per head group in bold.}
\label{tab:gq-body-tied-llama1b-instruct}
\begin{center}
\scriptsize
\renewcommand{\arraystretch}{0.9}
\setlength{\tabcolsep}{3pt}
\begin{tabular}{llrrrllll}
\multicolumn{9}{c}{\bf Llama-3.2-1B-Instruct} \\ \hline
\multicolumn{1}{c}{\rule{0pt}{2.6ex}\bf Body} & \multicolumn{1}{c}{\bf Method} & \multicolumn{1}{c}{\bf Head} & \multicolumn{1}{c}{\bf Size (MiB)} & \multicolumn{1}{c}{\bf $\Delta$size} & \multicolumn{1}{c}{\bf WT2 PPL ($\downarrow$)} & \multicolumn{1}{c}{\bf WT2 KL ($\downarrow$)} & \multicolumn{1}{c}{\bf C4 PPL ($\downarrow$)} & \multicolumn{1}{c}{\bf C4 KL ($\downarrow$)}
\\ \hline
\rule{0pt}{2.6ex}2 & fp16 & 16 & 737.3 & 0.0\% & 26.81 & 0.497 & 38.54 & 0.511 \\
\cline{2-9}
 & \rule{0pt}{2.6ex}WS & 2 & 297.0 & $-$59.7\% & 31.53 ($+$17.6\%) & 0.671 ($+$34.9\%) & 47.14 ($+$22.3\%) & 0.732 ($+$43.3\%) \\
 & \textbf{SW (ours)} & 2 & 297.0 & $-$59.7\% & \textbf{27.80 ($+$3.7\%)} & \textbf{0.536 ($+$7.9\%)} & \textbf{42.39 ($+$10.0\%)} & \textbf{0.607 ($+$18.8\%)} \\
\cline{2-9}
 & \rule{0pt}{2.6ex}WS & 3 & 327.7 & $-$55.6\% & 27.60 ($+$3.0\%) & 0.533 ($+$7.2\%) & 40.75 ($+$5.7\%) & 0.570 ($+$11.7\%) \\
 & \textbf{SW (ours)} & 3 & 327.7 & $-$55.6\% & \textbf{27.03 ($+$0.8\%)} & \textbf{0.506 ($+$1.7\%)} & \textbf{39.49 ($+$2.5\%)} & \textbf{0.536 ($+$4.9\%)} \\
\cline{2-9}
 & \rule{0pt}{2.6ex}WS & 4 & 358.4 & $-$51.4\% & 27.16 ($+$1.3\%) & 0.510 ($+$2.6\%) & 39.03 ($+$1.3\%) & 0.524 ($+$2.6\%) \\
 & \textbf{SW (ours)} & 4 & 358.4 & $-$51.4\% & \textbf{26.89 ($+$0.3\%)} & \textbf{0.501 ($+$0.7\%)} & \textbf{38.99 ($+$1.2\%)} & \textbf{0.521 ($+$2.1\%)} \\
\hline
\rule{0pt}{2.6ex}4 & fp16 & 16 & 962.6 & 0.0\% & 16.53 & 0.028 & 24.54 & 0.028 \\
\cline{2-9}
 & \rule{0pt}{2.6ex}WS & 2 & 532.5 & $-$44.7\% & 18.14 ($+$9.7\%) & 0.138 ($+$396.2\%) & 28.11 ($+$14.5\%) & 0.178 ($+$531.5\%) \\
 & \textbf{SW (ours)} & 2 & 532.5 & $-$44.7\% & \textbf{17.01 ($+$2.9\%)} & \textbf{0.056 ($+$101.9\%)} & \textbf{26.60 ($+$8.4\%)} & \textbf{0.106 ($+$276.9\%)} \\
\cline{2-9}
 & \rule{0pt}{2.6ex}WS & 3 & 563.2 & $-$41.5\% & 16.83 ($+$1.8\%) & 0.052 ($+$88.9\%) & 25.26 ($+$2.9\%) & 0.063 ($+$122.3\%) \\
 & \textbf{SW (ours)} & 3 & 563.2 & $-$41.5\% & \textbf{16.64 ($+$0.7\%)} & \textbf{0.034 ($+$24.1\%)} & \textbf{25.12 ($+$2.3\%)} & \textbf{0.052 ($+$84.6\%)} \\
\cline{2-9}
 & \rule{0pt}{2.6ex}WS & 4 & 594.0 & $-$38.3\% & 16.65 ($+$0.7\%) & 0.036 ($+$28.0\%) & \textbf{24.70 ($+$0.6\%)} & 0.037 ($+$32.4\%) \\
 & \textbf{SW (ours)} & 4 & 594.0 & $-$38.3\% & \textbf{16.58 ($+$0.3\%)} & \textbf{0.030 ($+$9.3\%)} & 24.77 ($+$0.9\%) & \textbf{0.035 ($+$25.3\%)} \\
\hline
\end{tabular}
\end{center}
\end{table}

Table~\ref{tab:gq-body-tied-llama1b-instruct} shows two of the three body rates; Table~\ref{tab:gq-body-tied-llama1b-instruct-full} gives the third.
The head is $21.3\%$ of the fp16 model's parameters, but once the body is quantized it holds $52\%$ to $68\%$ of the stored bytes, so quantizing it is the only remaining way to shrink the model.
A 2-bit SoftWater head removes $45$--$60\%$ of the stored bytes for a $2.9$--$3.7\%$ WT2 perplexity increase, against $9.7$--$17.6\%$ for WaterSIC, and SoftWater is better in $35$ of $36$ cells; at 4 bits the head is essentially free, costing $0.2$--$0.3\%$ while still removing at least $38\%$.
The head also dominates as bodies improve: on the 4-bit body a 2-bit head already doubles total KL under SoftWater and quintuples it under the released quantizer.

\subsection{Domain-targeted calibration}
\label{sub:domain-targeted}

Rate allocation varies across domains, so we target calibration to five themes, EN Wikipedia, DE Wikipedia, Python, Math web and Case law, and evaluate in and out of domain; Table~\ref{tab:crossover-corpora} in the appendix gives the corpora, splits and fields.
We calibrate on $1.05$M tokens rather than $131$k, since at $131$k most of the vocabulary is never observed and the domains would not be distinguishable through $\tilde{\lambda}$.

\begin{table}
\caption{In-domain next-token top-1 accuracy of Llama-3.2-1B-Instruct, fp16 body with a tied head quantized at 2, 3 and 4 bits.
Each arm is calibrated on the domain it is evaluated on.
KL: divergence from the all-fp16 model on the matching stream.
WS: WaterSIC w/o FT.
Best value per rate in bold.}
\label{tab:crossover-llama1b-instruct-diag}
\begin{center}
\scriptsize
\renewcommand{\arraystretch}{0.9}
\setlength{\tabcolsep}{3pt}
\begin{tabular}{llrrrrrrrrrr}
& & \multicolumn{5}{c}{\bf Top-1 accuracy ($\%$, $\uparrow$)} & \multicolumn{5}{c}{\bf KL ($\downarrow$)} \\
\cmidrule(lr){3-7}\cmidrule(lr){8-12}
\multicolumn{1}{c}{\rule{0pt}{2.6ex}\bf Head} & \multicolumn{1}{c}{\bf Method} & \multicolumn{1}{c}{\bf WT2} & \multicolumn{1}{c}{\bf DeWiki} & \multicolumn{1}{c}{\bf Code} & \multicolumn{1}{c}{\bf Math} & \multicolumn{1}{c}{\bf Law} & \multicolumn{1}{c}{\bf WT2} & \multicolumn{1}{c}{\bf DeWiki} & \multicolumn{1}{c}{\bf Code} & \multicolumn{1}{c}{\bf Math} & \multicolumn{1}{c}{\bf Law}
\\ \hline
\rule{0pt}{2.6ex}16 & fp16 & 45.6 & 52.9 & 66.9 & 54.6 & 44.5 & 0.000 & 0.000 & 0.000 & 0.000 & 0.000 \\
\hline
\rule{0pt}{2.6ex}2 & WS & 44.7 & 50.0 & 65.6 & 53.5 & 42.7 & 0.096 & 0.183 & 0.099 & 0.090 & 0.163 \\
 & \textbf{SW (ours)} & \textbf{45.4} & \textbf{52.5} & \textbf{66.3} & \textbf{54.3} & \textbf{44.1} & \textbf{0.021} & \textbf{0.028} & \textbf{0.052} & \textbf{0.028} & \textbf{0.025} \\
\cline{1-12}
\rule{0pt}{2.6ex}3 & WS & 45.4 & 52.3 & 66.7 & 54.3 & 44.1 & 0.023 & 0.040 & 0.024 & 0.023 & 0.034 \\
 & \textbf{SW (ours)} & \textbf{45.5} & \textbf{52.6} & \textbf{66.8} & \textbf{54.5} & \textbf{44.4} & \textbf{0.005} & \textbf{0.011} & \textbf{0.014} & \textbf{0.009} & \textbf{0.007} \\
\cline{1-12}
\rule{0pt}{2.6ex}4 & WS & \textbf{45.6} & 52.6 & 66.8 & \textbf{54.5} & 44.3 & 0.006 & 0.012 & 0.007 & 0.007 & 0.009 \\
 & \textbf{SW (ours)} & \textbf{45.6} & \textbf{52.9} & \textbf{66.9} & 54.4 & \textbf{44.5} & \textbf{0.002} & \textbf{0.003} & \textbf{0.004} & \textbf{0.002} & \textbf{0.002} \\
\hline
\end{tabular}
\end{center}
\end{table}

Table~\ref{tab:crossover-llama1b-instruct-diag} shows the in-domain results from Table~\ref{tab:crossover-llama1b-instruct}.
The targeting claim holds exactly on KL: for every evaluation domain and head rate, the lowest KL among the ten arms is the SoftWater head calibrated on that domain, and at 2 bits matching calibration lowers KL by $2.0\times$ to $5.0\times$ over the best mismatched arm.
On the diagonal SoftWater also outperforms the released quantizer everywhere at 2 bits, by $1.9\times$ to $6.5\times$ in KL, and wins $41$ of $50$ stream cells overall; seven of the nine losses are the code column under non-code calibration, the expected cost of sharpening allocation toward calibration-frequent tokens.

\subsection{Zero-shot tasks}
\label{sub:zeroshot}

Perplexity and KL measure exactly what SoftWater controls, so we now check downstream accuracy.
We reuse the configurations of \S\ref{sub:quantized-body} with head arms bit-identical to the perplexity runs, and evaluate zero-shot with lm-evaluation-harness 0.4.12 on ARC-Easy, ARC-Challenge, HellaSwag, LAMBADA, WinoGrande, PIQA and OpenBookQA, reporting normalized accuracy where defined and accuracy for LAMBADA.

\begin{table}
\caption{Zero-shot accuracy (higher better) of Llama-3.2-1B-Instruct with a \textbf{2-bit tied head} on GuidedQuant bodies at 2, 3 and 4 bits.
WS: WaterSIC w/o FT;
SW: SoftWater.
CTX: context length of $131$k WT2-train tokens.
lm-evaluation-harness 0.4.12, 0-shot.
Best value per body block in bold.}
\label{tab:zeroshot-llama1b-instruct}
\begin{center}
\scriptsize
\renewcommand{\arraystretch}{0.9}
\setlength{\tabcolsep}{3pt}
\begin{tabular}{llrrrrrrrrr}
\multicolumn{1}{c}{\rule{0pt}{2.6ex}\bf Body} & \multicolumn{1}{c}{\bf Method} & \multicolumn{1}{c}{\bf CTX} & \multicolumn{1}{c}{\bf ARC-e} & \multicolumn{1}{c}{\bf ARC-c} & \multicolumn{1}{c}{\bf HellaS} & \multicolumn{1}{c}{\bf LAMB} & \multicolumn{1}{c}{\bf Wino} & \multicolumn{1}{c}{\bf PIQA} & \multicolumn{1}{c}{\bf OBQA} & \multicolumn{1}{c}{\bf Mean}
\\ \hline
\rule{0pt}{2.6ex}16
 & fp16 head        & ---  & 63.8 & 37.8 & 61.7 & 60.1 & 61.6 & 74.9 & 37.2 & 56.72 \\
\hline
\rule{0pt}{2.6ex}2
 & fp16 head        & ---  & 58.0 & 32.4 & 49.3 & 46.3 & 55.5 & 69.2 & 31.4 & 48.86 \\
\cline{2-11}
 & \rule{0pt}{2.6ex}WS & 128  & 53.6 & 31.1 & 47.2 & 37.3 & 55.7 & 67.2 & 30.6 & 46.11 \\
 & WS               & 1024 & 53.2 & 31.5 & 47.4 & 36.8 & 55.3 & 67.7 & 29.4 & 45.89 \\
 & \textbf{SW (ours)} & 128  & \bf 56.5 & \bf 31.7 & \bf 48.7 & \bf 45.4 & \bf 56.8 & 67.8 & \bf 32.6 & \bf 48.49 \\
 & \textbf{SW (ours)} & 1024 & 54.7 & 31.1 & 47.9 & 43.9 & 55.5 & \bf 68.5 & 31.8 & 47.63 \\
\hline
\rule{0pt}{2.6ex}3
 & fp16 head        & ---  & 63.0 & 36.3 & 58.3 & 59.5 & 61.7 & 73.1 & 35.6 & 55.35 \\
\cline{2-11}
 & \rule{0pt}{2.6ex}WS & 128  & 59.7 & 34.5 & 57.0 & 50.7 & \bf 60.5 & 72.2 & 35.4 & 52.84 \\
 & WS               & 1024 & 57.9 & 32.8 & 57.2 & 53.1 & \bf 60.5 & 71.8 & 35.6 & 52.68 \\
 & \textbf{SW (ours)} & 128  & \bf 62.8 & \bf 36.6 & 57.2 & \bf 58.5 & 60.2 & \bf 73.0 & 35.2 & \bf 54.78 \\
 & \textbf{SW (ours)} & 1024 & 61.0 & 35.4 & \bf 57.5 & 58.4 & 60.2 & 72.3 & \bf 36.2 & 54.43 \\
\hline
\rule{0pt}{2.6ex}4
 & fp16 head        & ---  & 62.7 & 37.7 & 60.3 & 60.1 & 62.0 & 73.6 & 37.8 & 56.30 \\
\cline{2-11}
 & \rule{0pt}{2.6ex}WS & 128  & 60.5 & 36.6 & 58.9 & 53.6 & 61.5 & 72.2 & 35.8 & 54.16 \\
 & WS               & 1024 & 60.3 & 37.0 & 59.2 & 54.0 & 60.9 & 72.1 & \bf 37.2 & 54.39 \\
 & \textbf{SW (ours)} & 128  & \bf 62.2 & 37.4 & \bf 59.6 & \bf 59.4 & \bf 61.8 & \bf 73.2 & 36.4 & \bf 55.72 \\
 & \textbf{SW (ours)} & 1024 & 61.2 & \bf 37.5 & 59.3 & 59.2 & 61.6 & 72.7 & 35.0 & 55.23 \\
\hline
\end{tabular}
\end{center}
\end{table}

Table~\ref{tab:zeroshot-llama1b-instruct} shows 2-bit heads on all three bodies; Tables~\ref{tab:zeroshot-llama1b-instruct-b2}--\ref{tab:zeroshot-llama1b-instruct-b4} give every head rate.
At a 2-bit head SoftWater roughly halves the mean accuracy lost to head quantization on every body, cutting the drop from $2.97$, $2.67$ and $1.91$ points to $1.23$, $0.92$ and $1.07$; the gap narrows with head rate and closes at 4 bits.
The difference is concentrated in LAMBADA, the only task scored by next-token prediction rather than by ranking continuations, and so the one KL controls directly: at 2 bits SoftWater gains $5.2$ to $7.1$ points.
Sharpening the grids of frequent, low-variance classes protects exactly the token-level distribution KL measures.

\paragraph{Calibration diversity.}
Unlike $\Sigma_{X}$, the extra per-class statistic is a token frequency, so it benefits from many distinct contexts rather than many tokens.
Holding the budget at $131$k tokens and re-chunking into $128$-token contexts instead of $1024$ lifts the 2-bit-head mean by $0.35$ to $0.86$ points, while moving WaterSIC by at most $0.23$; quadrupling the budget at context length $1024$ helps by a similar amount and buys nothing that re-chunking does not.
Context count, not token count, is the binding resource for $\tilde{\lambda}$.

\subsection{Class-side statistic ablation}
\label{sub:class-side-statistics}

For most tokens $\tilde{p}_{k}(1-\tilde{p}_{k})\approx \tilde{p}_{k}$ because $\tilde{p}_{k}$ is small, so we ask whether the second-moment correction matters: we replace the average softmax curvature $\E_{X}[\tilde{p}_{k}(1-\tilde{p}_{k})]$ with the average marginal $\bar{p}_{k}=\E_{X}[\tilde{p}_{k}]$.
We also replace it with $\tilde{\pi}_{k}$, the \emph{unigram frequency} of token $k$ smoothed the same way, which requires no forward pass and asks whether the model carries information the corpus does not.

Table~\ref{tab:class-side-statistics} shows the results.
SoftWater and the $\bar{p}_{k}$-variant are close in domain, but SoftWater wins every out-of-domain KL cell and $12$ of $15$ in-domain ones, so the second-moment correction is small but systematic and is what makes the allocation transfer off the calibration domain.
Both variants beat $\tilde{\pi}_{k}$ at every rate on every model, by roughly $2\times$ in KL, so the model's own marginal carries information the corpus count does not.
Since $\bar{\lambda}_{k}$ and $\bar{p}_{k}$ come from the same forward pass, SoftWater is the best of the three.

\subsection{Separability}
\label{sub:separability-exp}

To test Assumption \ref{ass:separability}, define $Q_{k}=\frac{1}{2}\Delta_{k} \E_{X}[\lambda_{k}(X)XX^{\top}]\Delta_{k}^{\top}$ and $S_{k}=\frac{1}{2}\tilde{\lambda}_{k} \Delta_{k}\Sigma_{X}\Delta_{k}^{\top}$, the contribution of class $k$ to the true and approximate distortion, and the relative proxy error
\begin{align}
    \rho = \frac{\sum_{k=1}^{K}Q_{k}}{\sum_{k=1}^{K}S_{k}}-1,
    \label{def:relative-proxy-error}
\end{align}
whose rate penalty $\frac{1}{2}\log_{2}(1+\rho)$ is the extra bits needed to close the gap.
Since $S_{k}$ uses the smoothed $\tilde{\lambda}_{k}$ that the deployed scheme optimizes, $\rho$ combines the separability gap with the deliberate bias from smoothing.
Quantizing the head with SoftWater calibrated on WT2 and computing both sums on WT2 and C4, Table~\ref{tab:separability-ablation} shows the proxy underestimates the true distortion by at most 10\%, a rate penalty of at most $0.07$ bits.

Since the fit is imperfect, we ask whether better Kronecker factorizations exist.
We adapt the power iteration of \citet{tseng_model-preserving_2026} (Appendix~\ref{app:hessian-sketching}) to obtain sketches $a\otimes B$ with $a$ diagonal, run on $\E_{X}[\tilde{\lambda}(X)\otimes XX^{\top}]$ at $\epsilon = 0.1$, so its fixed point is the Frobenius-optimal factorization of the weight SoftWater targets.
Comparing three initializations, $(\tilde{\lambda}, \Sigma_{X})$ (SoftWater), $(I, \Sigma_{X})$ (WaterSIC) and $(I, I)$, Table~\ref{tab:sw-fixed-point} shows the SoftWater factors are close to optimal from the start.

\begin{table}
\caption{Separability diagnostics on Llama-3.2-1B.}
\label{tab:separability-and-fixed-point}
\begin{center}
\scriptsize
\renewcommand{\arraystretch}{0.9}
\setlength{\tabcolsep}{3pt}
\begin{subtable}[b]{0.40\linewidth}
\centering
\caption{$\rho$: relative proxy error \eqref{def:relative-proxy-error}.
bits $= \tfrac{1}{2}\log_2(1+\rho)$.
SW calibrated with WT2; 
Eval is the sample the ratio is measured on.}
\label{tab:separability-ablation}
\begin{tabular}{lrrrcrrr}
\multicolumn{1}{c}{\rule{0pt}{2.6ex}\bf Eval} & \multicolumn{3}{c}{WT2} & & \multicolumn{3}{c}{C4} \\
\cline{2-4} \cline{6-8}
\multicolumn{1}{c}{\rule{0pt}{2.6ex}\bf Rate} & 2 & 3 & 4 & & 2 & 3 & 4 \\
\hline
\rule{0pt}{2.6ex}
$\rho$ & $+$0.10 & $+$0.09 & $+$0.09 & & $+$0.08 & $+$0.08 & $+$0.07 \\
Bits & 0.07 & 0.06 & 0.06 & & 0.06 & 0.06 & 0.05 \\
\hline
\end{tabular}
\end{subtable}
\hfill
\begin{subtable}[b]{0.58\linewidth}
\centering
\caption{Convergence of alternating power iteration to its fixed point $(a^{\star}, B^{\star})$ from three initializations.
SW: $a_{0} = \tilde{\lambda}$, $B_{0} = \Sigma_{X}$; WS: $a_{0} = I$, $B_{0} = \Sigma_{X}$; Id: $a_{0} = I$, $B_{0} = I$.}
\label{tab:sw-fixed-point}
\begin{tabular}{lrrrcrrrcrrr}
\multicolumn{1}{c}{\rule{0pt}{2.6ex}\bf Init} & \multicolumn{3}{c}{\bf SW (ours)} & & \multicolumn{3}{c}{WS} & & \multicolumn{3}{c}{Id} \\
\cline{2-4} \cline{6-8} \cline{10-12}
\multicolumn{1}{c}{\rule{0pt}{2.6ex}\bf Round} & \textbf{0} & \textbf{1} & \textbf{2} & & 0 & 1 & 2 & & 0 & 1 & 2 \\
\hline
\rule{0pt}{2.6ex}
cos($a$, $a^{\star}$) & \textbf{0.99} & \textbf{1.00} & \textbf{1.00} & & 0.03 & 1.00 & 1.00 & & 0.03 & 0.99 & 1.00 \\
cos($B$, $B^{\star}$) & \textbf{0.98} & \textbf{1.00} & \textbf{1.00} & & 0.98 & 1.00 & 1.00 & & 0.05 & 1.00 & 1.00 \\
\hline
\end{tabular}
\end{subtable}
\end{center}
\end{table}

\section{Conclusion}
\label{sec:conclusion}
We posed softmax-layer quantization as a rate-distortion problem under output KL, and reduced it via a separable second-order surrogate to a class-rescaled lattice whose two statistics come from one calibration pass.
SoftWater allocates rate by class frequency and variance, with a smoothing prior that caps the spread on uncovered classes and interpolates back to WaterSIC as the prior takes over.
Across five models from 1B to 32B parameters it outperforms the released WaterSIC quantizer at matched head rates without any of that pipeline's refinements, halves the zero-shot accuracy lost to a 2-bit head, targets a deployment domain through its calibration statistic, and shrinks whole-model size by up to $60\%$.
Two limitations bound these results: the separable surrogate is accurate to within $10\%$ on the models we measure but not guaranteed in general, and the deployed-model experiments rest on a single family of released quantized bodies.
Beyond LLM heads, the analysis applies to any linear-softmax layer with a fixed class dimension, including vision and speech classifiers and MoE routers, whose skewed expert utilization mirrors the Zipfian structure we exploit.

\clearpage

\bibliography{AI,NA}
\bibliographystyle{iclr2026_conference}

\clearpage
\appendix

\section{Experiment details}

\subsection{Head-only isolation}

\begin{table}[H]
\caption{Head-only isolation: the body is kept in fp16 and only the head matrix is quantized; tied models (Llama-3.2-1B, Qwen3-1.7B) retain an fp16 copy of the embedding.
KL is measured against the all-fp16 model on identical token streams.
WS: WaterSIC w/o FT.
Parentheses after each model name give its vocabulary size $V$; on PPL cells, the increase over the fp16 row of the same model; on SW KL cells, the WS/SW KL ratio at that rate.
Best value per head group in bold.}
\label{tab:head-only-fp16-body-full}
\begin{center}
\scriptsize
\renewcommand{\arraystretch}{0.9}
\setlength{\tabcolsep}{3pt}
\begin{tabular}{lrllll}
\multicolumn{1}{c}{\bf Method} & \multicolumn{1}{c}{\bf Head} & \multicolumn{1}{c}{\bf WT2 PPL ($\downarrow$)} & \multicolumn{1}{c}{\bf WT2 KL ($\downarrow$)} & \multicolumn{1}{c}{\bf C4 PPL ($\downarrow$)} & \multicolumn{1}{c}{\bf C4 KL ($\downarrow$)}
\\ \hline
\multicolumn{6}{c}{\rule{0pt}{2.6ex}\bf Llama-3.2-1B (128{,}256)} \\ \hline
\rule{0pt}{2.6ex}fp16 & 16 & 11.313 & 0 & 15.614 & 0 \\
\cline{1-6}
\rule{0pt}{2.6ex}WS & 2 & 12.049 ($+$6.5\%) & 0.06192 & 17.283 ($+$10.7\%) & 0.09957 \\
\textbf{SW (ours)} & 2 & \textbf{11.419 ($+$0.9\%)} & \textbf{0.00957 (6.5$\times$)} & \textbf{16.035 ($+$2.7\%)} & \textbf{0.02834 (3.5$\times$)} \\
\cline{1-6}
\rule{0pt}{2.6ex}WS & 3 & 11.469 ($+$1.4\%) & 0.01364 & 15.965 ($+$2.2\%) & 0.02176 \\
\textbf{SW (ours)} & 3 & \textbf{11.342 ($+$0.3\%)} & \textbf{0.00206 (6.6$\times$)} & \textbf{15.704 ($+$0.6\%)} & \textbf{0.00572 (3.8$\times$)} \\
\cline{1-6}
\rule{0pt}{2.6ex}WS & 4 & 11.356 ($+$0.4\%) & 0.00319 & 15.713 ($+$0.6\%) & 0.00526 \\
\textbf{SW (ours)} & 4 & \textbf{11.320 ($+$0.1\%)} & \textbf{0.00047 (6.8$\times$)} & \textbf{15.640 ($+$0.2\%)} & \textbf{0.00138 (3.8$\times$)} \\
\hline
\multicolumn{6}{c}{\rule{0pt}{3.2ex}\bf Qwen3-1.7B (151{,}936)} \\ \hline
\rule{0pt}{2.6ex}fp16 & 16 & 18.285 & 0 & 22.621 & 0 \\
\cline{1-6}
\rule{0pt}{2.6ex}WS & 2 & 18.876 ($+$3.2\%) & 0.06788 & 24.129 ($+$6.7\%) & 0.09054 \\
\textbf{SW (ours)} & 2 & \textbf{18.460 ($+$1.0\%)} & \textbf{0.00900 (7.5$\times$)} & \textbf{23.075 ($+$2.0\%)} & \textbf{0.01998 (4.5$\times$)} \\
\cline{1-6}
\rule{0pt}{2.6ex}WS & 3 & 18.374 ($+$0.5\%) & 0.01448 & 22.916 ($+$1.3\%) & 0.02017 \\
\textbf{SW (ours)} & 3 & \textbf{18.317 ($+$0.2\%)} & \textbf{0.00199 (7.3$\times$)} & \textbf{22.724 ($+$0.5\%)} & \textbf{0.00443 (4.6$\times$)} \\
\cline{1-6}
\rule{0pt}{2.6ex}WS & 4 & \textbf{18.277 ($-$0.0\%)} & 0.00347 & 22.650 ($+$0.1\%) & 0.00487 \\
\textbf{SW (ours)} & 4 & 18.298 ($+$0.1\%) & \textbf{0.00045 (7.7$\times$)} & \textbf{22.633 ($+$0.1\%)} & \textbf{0.00105 (4.6$\times$)} \\
\hline
\multicolumn{6}{c}{\rule{0pt}{3.2ex}\bf Llama-3.1-8B-Instruct (128{,}256)} \\ \hline
\rule{0pt}{2.6ex}fp16 & 16 & 8.388 & 0 & 14.189 & 0 \\
\cline{1-6}
\rule{0pt}{2.6ex}WS & 2 & 8.758 ($+$4.4\%) & 0.04583 & 15.462 ($+$9.0\%) & 0.07487 \\
\textbf{SW (ours)} & 2 & \textbf{8.437 ($+$0.6\%)} & \textbf{0.00636 (7.2$\times$)} & \textbf{14.534 ($+$2.4\%)} & \textbf{0.02140 (3.5$\times$)} \\
\cline{1-6}
\rule{0pt}{2.6ex}WS & 3 & 8.458 ($+$0.8\%) & 0.00940 & 14.439 ($+$1.8\%) & 0.01622 \\
\textbf{SW (ours)} & 3 & \textbf{8.398 ($+$0.1\%)} & \textbf{0.00135 (7.0$\times$)} & \textbf{14.258 ($+$0.5\%)} & \textbf{0.00437 (3.7$\times$)} \\
\cline{1-6}
\rule{0pt}{2.6ex}WS & 4 & 8.402 ($+$0.2\%) & 0.00221 & 14.255 ($+$0.5\%) & 0.00383 \\
\textbf{SW (ours)} & 4 & \textbf{8.391 ($+$0.0\%)} & \textbf{0.00032 (6.8$\times$)} & \textbf{14.205 ($+$0.1\%)} & \textbf{0.00103 (3.7$\times$)} \\
\hline
\multicolumn{6}{c}{\rule{0pt}{3.2ex}\bf Qwen3-8B (151{,}936)} \\ \hline
\rule{0pt}{2.6ex}fp16 & 16 & 10.832 & 0 & 15.486 & 0 \\
\cline{1-6}
\rule{0pt}{2.6ex}WS & 2 & 11.081 ($+$2.3\%) & 0.04592 & 16.229 ($+$4.8\%) & 0.06861 \\
\textbf{SW (ours)} & 2 & \textbf{10.898 ($+$0.6\%)} & \textbf{0.00654 (7.0$\times$)} & \textbf{15.689 ($+$1.3\%)} & \textbf{0.01385 (5.0$\times$)} \\
\cline{1-6}
\rule{0pt}{2.6ex}WS & 3 & 10.877 ($+$0.4\%) & 0.00964 & 15.572 ($+$0.6\%) & 0.01319 \\
\textbf{SW (ours)} & 3 & \textbf{10.848 ($+$0.1\%)} & \textbf{0.00132 (7.3$\times$)} & \textbf{15.523 ($+$0.2\%)} & \textbf{0.00295 (4.5$\times$)} \\
\cline{1-6}
\rule{0pt}{2.6ex}WS & 4 & 10.854 ($+$0.2\%) & 0.00230 & 15.511 ($+$0.2\%) & 0.00320 \\
\textbf{SW (ours)} & 4 & \textbf{10.836 ($+$0.0\%)} & \textbf{0.00031 (7.3$\times$)} & \textbf{15.501 ($+$0.1\%)} & \textbf{0.00073 (4.4$\times$)} \\
\hline
\multicolumn{6}{c}{\rule{0pt}{3.2ex}\bf Qwen3-32B (151{,}936)} \\ \hline
\rule{0pt}{2.6ex}fp16 & 16 & 8.452 & 0 & 12.269 & 0 \\
\cline{1-6}
\rule{0pt}{2.6ex}WS & 2 & 8.756 ($+$3.6\%) & 0.05835 & 13.029 ($+$6.2\%) & 0.08106 \\
\textbf{SW (ours)} & 2 & \textbf{8.498 ($+$0.5\%)} & \textbf{0.00703 (8.3$\times$)} & \textbf{12.479 ($+$1.7\%)} & \textbf{0.01596 (5.1$\times$)} \\
\cline{1-6}
\rule{0pt}{2.6ex}WS & 3 & 8.500 ($+$0.6\%) & 0.01246 & 12.439 ($+$1.4\%) & 0.01683 \\
\textbf{SW (ours)} & 3 & \textbf{8.464 ($+$0.1\%)} & \textbf{0.00147 (8.5$\times$)} & \textbf{12.315 ($+$0.4\%)} & \textbf{0.00341 (4.9$\times$)} \\
\cline{1-6}
\rule{0pt}{2.6ex}WS & 4 & 8.477 ($+$0.3\%) & 0.00295 & 12.302 ($+$0.3\%) & 0.00397 \\
\textbf{SW (ours)} & 4 & \textbf{8.457 ($+$0.1\%)} & \textbf{0.00035 (8.4$\times$)} & \textbf{12.275 ($+$0.1\%)} & \textbf{0.00081 (4.9$\times$)} \\
\hline
\end{tabular}
\end{center}
\end{table}

\subsection{Quantized body, tied head}

\begin{table}[H]
\caption{Perplexity (PPL) on WikiText2 (WT2) and C4, and KL divergence from the baseline model, consisting of the body quantized at 2, 3 and 4 bits with GuidedQuant (BlockLDLQ, QTIP format, released checkpoints) and the head kept in fp16.
Body and head columns give the respective rates.
The quantized head matrix is tied (also the input embedding).
WS: WaterSIC w/o FT.
Parentheses give the increase over the fp16-head row of the same body.
Whole-model size in MiB and reduction ($\Delta$size) after quantizing the head.
Best value per head group in bold.}
\label{tab:gq-body-tied-llama1b-instruct-full}
\begin{center}
\scriptsize
\renewcommand{\arraystretch}{0.9}
\setlength{\tabcolsep}{3pt}
\begin{tabular}{llrrrllll}
\multicolumn{9}{c}{\bf Llama-3.2-1B-Instruct} \\ \hline
\multicolumn{1}{c}{\rule{0pt}{2.6ex}\bf Body} & \multicolumn{1}{c}{\bf Method} & \multicolumn{1}{c}{\bf Head} & \multicolumn{1}{c}{\bf Size (MiB)} & \multicolumn{1}{c}{\bf $\Delta$size} & \multicolumn{1}{c}{\bf WT2 PPL ($\downarrow$)} & \multicolumn{1}{c}{\bf WT2 KL ($\downarrow$)} & \multicolumn{1}{c}{\bf C4 PPL ($\downarrow$)} & \multicolumn{1}{c}{\bf C4 KL ($\downarrow$)}
\\ \hline
\rule{0pt}{2.6ex}2.00 & fp16 & 16.00 & 737.3 & 0.0\% & 26.81 & 0.497 & 38.54 & 0.511 \\
\cline{2-9}
 & \rule{0pt}{2.6ex}WS & 2.00 & 297.0 & $-$59.7\% & 31.53 ($+$17.6\%) & 0.671 ($+$34.9\%) & 47.14 ($+$22.3\%) & 0.732 ($+$43.3\%) \\
 & \textbf{SW (ours)} & 2.00 & 297.0 & $-$59.7\% & \textbf{27.80 ($+$3.7\%)} & \textbf{0.536 ($+$7.9\%)} & \textbf{42.39 ($+$10.0\%)} & \textbf{0.607 ($+$18.8\%)} \\
\cline{2-9}
 & \rule{0pt}{2.6ex}WS & 3.00 & 327.7 & $-$55.6\% & 27.60 ($+$3.0\%) & 0.533 ($+$7.2\%) & 40.75 ($+$5.7\%) & 0.570 ($+$11.7\%) \\
 & \textbf{SW (ours)} & 3.00 & 327.7 & $-$55.6\% & \textbf{27.03 ($+$0.8\%)} & \textbf{0.506 ($+$1.7\%)} & \textbf{39.49 ($+$2.5\%)} & \textbf{0.536 ($+$4.9\%)} \\
\cline{2-9}
 & \rule{0pt}{2.6ex}WS & 4.00 & 358.4 & $-$51.4\% & 27.16 ($+$1.3\%) & 0.510 ($+$2.6\%) & 39.03 ($+$1.3\%) & 0.524 ($+$2.6\%) \\
 & \textbf{SW (ours)} & 4.00 & 358.4 & $-$51.4\% & \textbf{26.89 ($+$0.3\%)} & \textbf{0.501 ($+$0.7\%)} & \textbf{38.99 ($+$1.2\%)} & \textbf{0.521 ($+$2.1\%)} \\
\hline
\rule{0pt}{2.6ex}3.00 & fp16 & 16.00 & 850.0 & 0.0\% & 17.85 & 0.101 & 26.15 & 0.096 \\
\cline{2-9}
 & \rule{0pt}{2.6ex}WS & 2.00 & 409.6 & $-$51.8\% & 19.99 ($+$12.0\%) & 0.230 ($+$126.8\%) & 30.84 ($+$17.9\%) & 0.269 ($+$180.2\%) \\
 & \textbf{SW (ours)} & 2.00 & 409.6 & $-$51.8\% & \textbf{18.44 ($+$3.3\%)} & \textbf{0.136 ($+$34.5\%)} & \textbf{28.52 ($+$9.0\%)} & \textbf{0.180 ($+$87.5\%)} \\
\cline{2-9}
 & \rule{0pt}{2.6ex}WS & 3.00 & 440.3 & $-$48.2\% & 18.23 ($+$2.1\%) & 0.128 ($+$26.0\%) & 26.96 ($+$3.1\%) & 0.135 ($+$40.4\%) \\
 & \textbf{SW (ours)} & 3.00 & 440.3 & $-$48.2\% & \textbf{17.98 ($+$0.7\%)} & \textbf{0.109 ($+$7.1\%)} & \textbf{26.77 ($+$2.4\%)} & \textbf{0.119 ($+$23.8\%)} \\
\cline{2-9}
 & \rule{0pt}{2.6ex}WS & 4.00 & 471.0 & $-$44.6\% & 18.08 ($+$1.3\%) & 0.113 ($+$11.5\%) & 26.45 ($+$1.1\%) & 0.107 ($+$11.3\%) \\
 & \textbf{SW (ours)} & 4.00 & 471.0 & $-$44.6\% & \textbf{17.88 ($+$0.2\%)} & \textbf{0.104 ($+$2.3\%)} & \textbf{26.32 ($+$0.6\%)} & \textbf{0.102 ($+$6.7\%)} \\
\hline
\rule{0pt}{2.6ex}4.00 & fp16 & 16.00 & 962.6 & 0.0\% & 16.53 & 0.028 & 24.54 & 0.028 \\
\cline{2-9}
 & \rule{0pt}{2.6ex}WS & 2.00 & 532.5 & $-$44.7\% & 18.14 ($+$9.7\%) & 0.138 ($+$396.2\%) & 28.11 ($+$14.5\%) & 0.178 ($+$531.5\%) \\
 & \textbf{SW (ours)} & 2.00 & 532.5 & $-$44.7\% & \textbf{17.01 ($+$2.9\%)} & \textbf{0.056 ($+$101.9\%)} & \textbf{26.60 ($+$8.4\%)} & \textbf{0.106 ($+$276.9\%)} \\
\cline{2-9}
 & \rule{0pt}{2.6ex}WS & 3.00 & 563.2 & $-$41.5\% & 16.83 ($+$1.8\%) & 0.052 ($+$88.9\%) & 25.26 ($+$2.9\%) & 0.063 ($+$122.3\%) \\
 & \textbf{SW (ours)} & 3.00 & 563.2 & $-$41.5\% & \textbf{16.64 ($+$0.7\%)} & \textbf{0.034 ($+$24.1\%)} & \textbf{25.12 ($+$2.3\%)} & \textbf{0.052 ($+$84.6\%)} \\
\cline{2-9}
 & \rule{0pt}{2.6ex}WS & 4.00 & 594.0 & $-$38.3\% & 16.65 ($+$0.7\%) & 0.036 ($+$28.0\%) & \textbf{24.70 ($+$0.6\%)} & 0.037 ($+$32.4\%) \\
 & \textbf{SW (ours)} & 4.00 & 594.0 & $-$38.3\% & \textbf{16.58 ($+$0.3\%)} & \textbf{0.030 ($+$9.3\%)} & 24.77 ($+$0.9\%) & \textbf{0.035 ($+$25.3\%)} \\
\hline
\end{tabular}
\end{center}
\end{table}

\subsection{Domain-targeted calibration}

\begin{table}[H]
\caption{Domains used for calibration and evaluation.
Calibration and evaluation slices are disjoint: WT2 by split, the remaining domains by offsetting the calibration stream past the evaluation slice (skip 32).
Every calibration slice is $1024 \times 1024$ tokens and every evaluation slice $32 \times 1024$ tokens.}
\label{tab:crossover-corpora}
\begin{center}
\scriptsize
\renewcommand{\arraystretch}{0.9}
\setlength{\tabcolsep}{3pt}
\begin{tabular}{lllll}
\multicolumn{1}{c}{\bf Domain} & \multicolumn{1}{c}{\bf Corpus} & \multicolumn{1}{c}{\bf Config} & \multicolumn{1}{c}{\bf Split} & \multicolumn{1}{c}{\bf Field}
\\ \hline
\rule{0pt}{2.6ex}WT2 & \texttt{wikitext} & \texttt{wikitext-2-raw-v1} & \texttt{train\,/\,test} & \texttt{text} \\
DeWiki & \texttt{wikimedia/wikipedia} & \texttt{20231101.de} & \texttt{train} & \texttt{text} \\
Code & \texttt{codeparrot/codeparrot-clean-valid} & --- & \texttt{train} & \texttt{content} \\
Math & \texttt{open-web-math/open-web-math} & --- & \texttt{train} & \texttt{text} \\
Law & \texttt{HFforLegal/case-law} & --- & \texttt{us} & \texttt{document} \\
\hline
\end{tabular}
\end{center}
\end{table}

\begin{table}[H]
\caption{Calibration $\times$ evaluation domain crossover on Llama-3.2-1B-Instruct, fp16 body with a tied head quantized at 2, 3 and 4 bits.
Each block gives the calibration domain used for both $\tilde{\lambda}$ and $\Sigma_X$; diagonal cells (calibration domain matches evaluation domain) are the targeting claim.
Accuracy columns: next-token top-1 accuracy in percent, higher is better.
KL columns: divergence from the all-fp16 model on the matching stream, lower is better.
WS: WaterSIC w/o FT.
The fp16-head row is repeated per rate as the reference.
Best value per calibration block in bold; ties bold both.}
\label{tab:crossover-llama1b-instruct}
\begin{center}
\scriptsize
\renewcommand{\arraystretch}{0.9}
\setlength{\tabcolsep}{3pt}
\begin{tabular}{llrrrrrrrrrr}
& & \multicolumn{5}{c}{\bf Accuracy ($\%$, $\uparrow$)} & \multicolumn{5}{c}{\bf KL ($\downarrow$)} \\
\cmidrule(lr){3-7}\cmidrule(lr){8-12}
\multicolumn{1}{c}{\rule{0pt}{2.6ex}\bf Calib} & \multicolumn{1}{c}{\bf Method} & \multicolumn{1}{c}{\bf WT2} & \multicolumn{1}{c}{\bf DeWiki} & \multicolumn{1}{c}{\bf Code} & \multicolumn{1}{c}{\bf Math} & \multicolumn{1}{c}{\bf Law} & \multicolumn{1}{c}{\bf WT2} & \multicolumn{1}{c}{\bf DeWiki} & \multicolumn{1}{c}{\bf Code} & \multicolumn{1}{c}{\bf Math} & \multicolumn{1}{c}{\bf Law}
\\ \hline
\multicolumn{12}{c}{\rule{0pt}{2.6ex}\bf Head @ 2 bits} \\ \hline
\rule{0pt}{2.6ex}--- & fp16 & 45.6 & 52.9 & 66.9 & 54.6 & 44.5 & 0.000 & 0.000 & 0.000 & 0.000 & 0.000 \\
\cline{1-12}
\rule{0pt}{2.6ex}WT2 & WS & 44.7 & 50.0 & \textbf{64.8} & 52.8 & 42.9 & 0.096 & 0.212 & \textbf{0.151} & 0.123 & 0.157 \\
 & \textbf{SW (ours)} & \textbf{45.4} & \textbf{50.5} & 64.2 & \textbf{53.8} & \textbf{43.6} & \textbf{0.021} & \textbf{0.200} & 0.218 & \textbf{0.080} & \textbf{0.086} \\
\cline{1-12}
\rule{0pt}{2.6ex}DeWiki & WS & 43.3 & 50.0 & \textbf{64.5} & 52.1 & 41.6 & 0.172 & 0.183 & \textbf{0.174} & 0.163 & 0.252 \\
 & \textbf{SW (ours)} & \textbf{44.2} & \textbf{52.5} & 64.1 & \textbf{53.0} & \textbf{43.0} & \textbf{0.117} & \textbf{0.028} & 0.219 & \textbf{0.110} & \textbf{0.141} \\
\cline{1-12}
\rule{0pt}{2.6ex}Code & WS & 43.4 & \textbf{49.5} & 65.6 & 53.1 & 42.0 & 0.173 & \textbf{0.245} & 0.099 & 0.121 & 0.210 \\
 & \textbf{SW (ours)} & \textbf{43.9} & \textbf{49.5} & \textbf{66.3} & \textbf{53.9} & \textbf{43.1} & \textbf{0.135} & 0.272 & \textbf{0.052} & \textbf{0.056} & \textbf{0.138} \\
\cline{1-12}
\rule{0pt}{2.6ex}Math & WS & 44.1 & 50.2 & 65.4 & 53.5 & 42.7 & 0.135 & 0.194 & \textbf{0.118} & 0.090 & 0.165 \\
 & \textbf{SW (ours)} & \textbf{44.8} & \textbf{51.4} & \textbf{65.6} & \textbf{54.3} & \textbf{43.7} & \textbf{0.062} & \textbf{0.139} & 0.119 & \textbf{0.028} & \textbf{0.083} \\
\cline{1-12}
\rule{0pt}{2.6ex}Law & WS & 43.5 & 49.4 & \textbf{64.8} & 52.7 & 42.7 & 0.144 & 0.262 & \textbf{0.157} & 0.135 & 0.163 \\
 & \textbf{SW (ours)} & \textbf{44.9} & \textbf{50.2} & 64.5 & \textbf{54.0} & \textbf{44.1} & \textbf{0.060} & \textbf{0.215} & 0.182 & \textbf{0.056} & \textbf{0.025} \\
\hline
\multicolumn{12}{c}{\rule{0pt}{3.2ex}\bf Head @ 3 bits} \\ \hline
\rule{0pt}{2.6ex}--- & fp16 & 45.6 & 52.9 & 66.9 & 54.6 & 44.5 & 0.000 & 0.000 & 0.000 & 0.000 & 0.000 \\
\cline{1-12}
\rule{0pt}{2.6ex}WT2 & WS & 45.4 & \textbf{52.2} & \textbf{66.5} & 54.1 & 43.9 & 0.023 & 0.053 & \textbf{0.034} & 0.028 & 0.040 \\
 & \textbf{SW (ours)} & \textbf{45.5} & \textbf{52.2} & 66.4 & \textbf{54.4} & \textbf{44.2} & \textbf{0.005} & \textbf{0.052} & 0.049 & \textbf{0.015} & \textbf{0.022} \\
\cline{1-12}
\rule{0pt}{2.6ex}DeWiki & WS & 45.1 & 52.3 & \textbf{66.4} & 54.1 & \textbf{44.0} & 0.037 & 0.040 & \textbf{0.037} & 0.034 & 0.055 \\
 & \textbf{SW (ours)} & \textbf{45.2} & \textbf{52.6} & 66.3 & \textbf{54.3} & 43.9 & \textbf{0.026} & \textbf{0.011} & 0.045 & \textbf{0.024} & \textbf{0.033} \\
\cline{1-12}
\rule{0pt}{2.6ex}Code & WS & 45.0 & \textbf{52.2} & 66.7 & 54.2 & 43.9 & 0.039 & \textbf{0.059} & 0.024 & 0.030 & 0.050 \\
 & \textbf{SW (ours)} & \textbf{45.2} & 52.1 & \textbf{66.8} & \textbf{54.4} & \textbf{44.2} & \textbf{0.027} & 0.062 & \textbf{0.014} & \textbf{0.013} & \textbf{0.030} \\
\cline{1-12}
\rule{0pt}{2.6ex}Math & WS & \textbf{45.4} & \textbf{52.2} & 66.5 & 54.3 & 44.0 & 0.030 & 0.047 & \textbf{0.027} & 0.023 & 0.043 \\
 & \textbf{SW (ours)} & \textbf{45.4} & \textbf{52.2} & \textbf{66.6} & \textbf{54.5} & \textbf{44.2} & \textbf{0.016} & \textbf{0.041} & 0.030 & \textbf{0.009} & \textbf{0.021} \\
\cline{1-12}
\rule{0pt}{2.6ex}Law & WS & 45.1 & 52.0 & \textbf{66.4} & 54.0 & 44.1 & 0.034 & 0.057 & \textbf{0.035} & 0.031 & 0.034 \\
 & \textbf{SW (ours)} & \textbf{45.5} & \textbf{52.2} & \textbf{66.4} & \textbf{54.5} & \textbf{44.4} & \textbf{0.016} & \textbf{0.056} & 0.042 & \textbf{0.013} & \textbf{0.007} \\
\hline
\multicolumn{12}{c}{\rule{0pt}{3.2ex}\bf Head @ 4 bits} \\ \hline
\rule{0pt}{2.6ex}--- & fp16 & 45.6 & 52.9 & 66.9 & 54.6 & 44.5 & 0.000 & 0.000 & 0.000 & 0.000 & 0.000 \\
\cline{1-12}
\rule{0pt}{2.6ex}WT2 & WS & \textbf{45.6} & 52.6 & \textbf{66.8} & \textbf{54.5} & 44.1 & 0.006 & \textbf{0.015} & \textbf{0.011} & 0.008 & 0.011 \\
 & \textbf{SW (ours)} & \textbf{45.6} & \textbf{52.7} & \textbf{66.8} & \textbf{54.5} & \textbf{44.4} & \textbf{0.002} & 0.016 & 0.014 & \textbf{0.005} & \textbf{0.007} \\
\cline{1-12}
\rule{0pt}{2.6ex}DeWiki & WS & \textbf{45.5} & 52.6 & \textbf{66.9} & \textbf{54.5} & 44.2 & 0.010 & 0.012 & \textbf{0.011} & 0.010 & 0.014 \\
 & \textbf{SW (ours)} & \textbf{45.5} & \textbf{52.9} & 66.7 & \textbf{54.5} & \textbf{44.3} & \textbf{0.009} & \textbf{0.003} & 0.015 & \textbf{0.008} & \textbf{0.011} \\
\cline{1-12}
\rule{0pt}{2.6ex}Code & WS & 45.3 & 52.5 & 66.8 & \textbf{54.5} & 44.2 & 0.011 & \textbf{0.015} & 0.007 & 0.008 & 0.014 \\
 & \textbf{SW (ours)} & \textbf{45.5} & \textbf{52.6} & \textbf{66.9} & \textbf{54.5} & \textbf{44.4} & \textbf{0.008} & 0.021 & \textbf{0.004} & \textbf{0.004} & \textbf{0.009} \\
\cline{1-12}
\rule{0pt}{2.6ex}Math & WS & 45.4 & \textbf{52.7} & \textbf{66.9} & \textbf{54.5} & 44.3 & 0.008 & \textbf{0.013} & \textbf{0.008} & 0.007 & 0.013 \\
 & \textbf{SW (ours)} & \textbf{45.6} & \textbf{52.7} & 66.8 & 54.4 & \textbf{44.4} & \textbf{0.006} & \textbf{0.013} & 0.009 & \textbf{0.002} & \textbf{0.007} \\
\cline{1-12}
\rule{0pt}{2.6ex}Law & WS & \textbf{45.5} & 52.5 & \textbf{66.8} & 54.4 & 44.3 & 0.009 & \textbf{0.015} & \textbf{0.009} & 0.008 & 0.009 \\
 & \textbf{SW (ours)} & \textbf{45.5} & \textbf{52.7} & \textbf{66.8} & \textbf{54.5} & \textbf{44.5} & \textbf{0.005} & \textbf{0.015} & 0.011 & \textbf{0.004} & \textbf{0.002} \\
\hline
\end{tabular}
\end{center}
\end{table}

\subsection{Zero-shot tasks}

\begin{table}[H]
\caption{Zero-shot accuracy (higher better) of Llama-3.2-1B-Instruct with a \textbf{2-bit GuidedQuant body} and a tied head quantized at 2, 3 and 4 bits.
WS: WaterSIC w/o FT; 
SW: SoftWater.
CTX: context length of 131k WT2-train tokens. 
lm-evaluation-harness 0.4.12, 0-shot. 
Best value per body block in bold.}
\label{tab:zeroshot-llama1b-instruct-b2}
\begin{center}
\scriptsize
\renewcommand{\arraystretch}{0.9}
\setlength{\tabcolsep}{3pt}
\begin{tabular}{llrrrrrrrrrr}
\multicolumn{1}{c}{\rule{0pt}{2.6ex}\bf Head} & \multicolumn{1}{c}{\bf Method} & \multicolumn{1}{c}{\bf CTX} & \multicolumn{1}{c}{\bf Tokens} & \multicolumn{1}{c}{\bf ARC-e} & \multicolumn{1}{c}{\bf ARC-c} & \multicolumn{1}{c}{\bf HellaS} & \multicolumn{1}{c}{\bf LAMB} & \multicolumn{1}{c}{\bf Wino} & \multicolumn{1}{c}{\bf PIQA} & \multicolumn{1}{c}{\bf OBQA} & \multicolumn{1}{c}{\bf Mean}
\\ \hline
\rule{0pt}{2.6ex}16
 & fp16 body        & ---  & ---   & 63.8 & 37.8 & 61.7 & 60.1 & 61.6 & 74.9 & 37.2 & 56.72 \\
16
 & fp16 head        & ---  & ---   & 58.0 & 32.4 & 49.3 & 46.3 & 55.5 & 69.2 & 31.4 & 48.86 \\
\hline
\rule{0pt}{2.6ex}2
 & WS               & 128  & 131k  & 53.6 & 31.1 & 47.2 & 37.3 & 55.7 & 67.2 & 30.6 & 46.11 \\
 & WS               & 256  & 131k  & 52.5 & 29.9 & 47.0 & 36.6 & 56.2 & 67.4 & 30.8 & 45.78 \\
 & WS               & 1024 & 131k  & 53.2 & 31.5 & 47.4 & 36.8 & 55.3 & 67.7 & 29.4 & 45.89 \\
 & WS               & 1024 & 1.05M & 54.0 & 30.7 & 47.2 & 38.4 & 55.6 & 67.1 & 28.8 & 45.97 \\
 & \textbf{SW (ours)} & 128  & 131k  & \bf 56.5 & \bf 31.7 & \bf 48.7 & \bf 45.4 & \bf 56.8 & 67.8 & \bf 32.6 & \bf 48.49 \\
 & \textbf{SW (ours)} & 256  & 131k  & 53.4 & 29.7 & 48.1 & 44.9 & 56.4 & 66.9 & 31.6 & 47.30 \\
 & \textbf{SW (ours)} & 1024 & 131k  & 54.7 & 31.1 & 47.9 & 43.9 & 55.5 & \bf 68.5 & 31.8 & 47.63 \\
 & \textbf{SW (ours)} & 1024 & 1.05M & 55.7 & \bf 31.7 & 48.3 & 44.8 & 56.7 & 68.0 & 30.0 & 47.87 \\
\cline{1-12}
\rule{0pt}{2.6ex}3
 & WS               & 128  & 131k  & 55.1 & 31.3 & 49.1 & 43.7 & 55.8 & 68.8 & 31.0 & 47.83 \\
 & WS               & 256  & 131k  & 55.2 & 31.3 & 48.9 & 43.8 & 55.2 & \bf 69.2 & 30.4 & 47.71 \\
 & WS               & 1024 & 131k  & 56.1 & 31.1 & 48.8 & 43.1 & 56.9 & 67.9 & 30.2 & 47.73 \\
 & WS               & 1024 & 1.05M & 56.3 & 31.6 & 49.1 & 43.6 & 56.5 & 68.0 & 30.0 & 47.87 \\
 & \textbf{SW (ours)} & 128  & 131k  & 57.4 & 32.1 & \bf 49.4 & 45.9 & 56.7 & 68.7 & \bf 31.6 & 48.82 \\
 & \textbf{SW (ours)} & 256  & 131k  & \bf 58.0 & 32.2 & 49.3 & \bf 46.7 & \bf 57.5 & 68.5 & 31.2 & \bf 49.04 \\
 & \textbf{SW (ours)} & 1024 & 131k  & 57.0 & \bf 32.5 & 49.0 & 45.6 & 56.6 & 68.8 & \bf 31.6 & 48.72 \\
 & \textbf{SW (ours)} & 1024 & 1.05M & 57.3 & 32.3 & 49.2 & 46.2 & 57.2 & 68.7 & \bf 31.6 & 48.92 \\
\cline{1-12}
\rule{0pt}{2.6ex}4
 & WS               & 128  & 131k  & 57.6 & 32.3 & 49.3 & 45.2 & 56.0 & 69.4 & 31.8 & 48.79 \\
 & WS               & 256  & 131k  & 57.7 & 32.3 & 49.4 & 45.2 & 55.6 & \bf 69.6 & 32.0 & 48.82 \\
 & WS               & 1024 & 131k  & 57.4 & 31.8 & \bf 49.5 & 45.5 & 55.7 & 69.2 & 32.0 & 48.74 \\
 & WS               & 1024 & 1.05M & \bf 58.1 & 32.1 & 49.4 & 45.6 & 55.9 & 68.9 & 32.0 & 48.85 \\
 & \textbf{SW (ours)} & 128  & 131k  & 57.4 & \bf 32.6 & 49.2 & 46.0 & \bf 56.4 & 68.8 & 31.6 & \bf 48.86 \\
 & \textbf{SW (ours)} & 256  & 131k  & 56.8 & 32.3 & 49.3 & 46.1 & 55.8 & 69.0 & 31.6 & 48.70 \\
 & \textbf{SW (ours)} & 1024 & 131k  & 57.3 & 31.7 & 49.2 & \bf 46.3 & 56.2 & 69.3 & 31.8 & 48.83 \\
 & \textbf{SW (ours)} & 1024 & 1.05M & 57.0 & 31.6 & 49.2 & 46.1 & 56.0 & 69.3 & \bf 32.2 & 48.77 \\
\hline
\end{tabular}
\end{center}
\end{table}

\begin{table}[H]
\caption{Zero-shot accuracy (higher better) of Llama-3.2-1B-Instruct with a \textbf{3-bit GuidedQuant body} and a tied head quantized at 2, 3 and 4 bits.
Details as in Table~\ref{tab:zeroshot-llama1b-instruct-b2}.
Best value per head group in bold.}
\label{tab:zeroshot-llama1b-instruct-b3}
\begin{center}
\scriptsize
\renewcommand{\arraystretch}{0.9}
\setlength{\tabcolsep}{3pt}
\begin{tabular}{llrrrrrrrr}
\multicolumn{1}{c}{\rule{0pt}{2.6ex}\bf Head} & \multicolumn{1}{c}{\bf Method} & \multicolumn{1}{c}{\bf ARC-e} & \multicolumn{1}{c}{\bf ARC-c} & \multicolumn{1}{c}{\bf HellaS} & \multicolumn{1}{c}{\bf LAMB} & \multicolumn{1}{c}{\bf Wino} & \multicolumn{1}{c}{\bf PIQA} & \multicolumn{1}{c}{\bf OBQA} & \multicolumn{1}{c}{\bf Mean}
\\ \hline
\rule{0pt}{2.6ex}16
 & fp16 body                  & 63.8 & 37.8 & 61.7 & 60.1 & 61.6 & 74.9 & 37.2 & 56.72 \\
16
 & fp16 head                  & 63.0 & 36.3 & 58.3 & 59.5 & 61.7 & 73.1 & 35.6 & 55.35 \\
\hline
\rule{0pt}{2.6ex}2
 & WS-c128                    & 59.7 & 34.5 & 57.0 & 50.7 & 60.5 & 72.2 & 35.4 & 52.84 \\
 & WS-c256                    & 59.6 & 34.6 & 56.8 & 51.6 & 59.5 & 71.4 & \bf 37.0 & 52.94 \\
 & WS-c1024                   & 57.9 & 32.8 & 57.2 & 53.1 & 60.5 & 71.8 & 35.6 & 52.68 \\
 & WS-1M                      & 59.4 & 34.0 & 56.8 & 52.3 & 59.7 & 71.7 & 36.6 & 52.93 \\
 & \textbf{SW-c128}           & \bf 62.8 & 36.6 & 57.2 & 58.5 & 60.2 & \bf 73.0 & 35.2 & 54.78 \\
 & \textbf{SW-c256}           & 62.5 & 36.6 & 57.4 & 58.5 & 60.2 & 72.3 & 36.4 & 54.85 \\
 & \textbf{SW-c1024}          & 61.0 & 35.4 & \bf 57.5 & 58.4 & 60.2 & 72.3 & 36.2 & 54.43 \\
 & \textbf{SW-1M}             & 61.3 & \bf 36.7 & 57.3 & \bf 58.7 & \bf 61.1 & 72.6 & 36.4 & \bf 54.87 \\
\cline{1-10}
\rule{0pt}{2.6ex}3
 & WS-c128                    & 63.0 & 36.4 & 57.7 & 58.3 & \bf 62.4 & \bf 73.4 & 36.0 & \bf 55.31 \\
 & WS-c256                    & \bf 63.1 & \bf 37.0 & 57.9 & 57.1 & 61.3 & 73.3 & 35.8 & 55.08 \\
 & WS-c1024                   & 61.8 & 36.8 & 58.0 & 57.9 & 61.1 & 73.0 & 35.4 & 54.84 \\
 & WS-1M                      & 62.1 & 36.5 & 58.0 & 58.5 & 61.1 & 72.9 & 35.6 & 54.95 \\
 & \textbf{SW-c128}           & 62.1 & 35.2 & \bf 58.3 & \bf 59.3 & 61.4 & 72.7 & \bf 37.2 & 55.18 \\
 & \textbf{SW-c256}           & 62.5 & 35.8 & \bf 58.3 & \bf 59.3 & 60.9 & 72.7 & 36.8 & 55.20 \\
 & \textbf{SW-c1024}          & 62.1 & 36.2 & 58.0 & 59.0 & 61.2 & 73.1 & 35.8 & 55.07 \\
 & \textbf{SW-1M}             & 62.2 & 36.2 & 58.1 & \bf 59.3 & 60.6 & 73.1 & 35.8 & 55.05 \\
\cline{1-10}
\rule{0pt}{2.6ex}4
 & WS-c128                    & 62.7 & 36.5 & \bf 58.3 & 58.8 & 60.9 & 73.4 & \bf 37.0 & 55.38 \\
 & WS-c256                    & 62.8 & 36.3 & 58.2 & 59.1 & \bf 61.8 & 73.4 & 36.4 & 55.43 \\
 & WS-c1024                   & \bf 63.1 & 36.1 & 58.2 & 59.0 & 61.6 & 73.4 & 35.8 & 55.31 \\
 & WS-1M                      & 63.0 & 36.6 & 58.1 & 58.8 & 60.9 & 72.9 & 35.8 & 55.16 \\
 & \textbf{SW-c128}           & \bf 63.1 & 36.0 & 58.2 & 59.6 & 61.1 & 72.7 & 35.6 & 55.17 \\
 & \textbf{SW-c256}           & 62.7 & 36.6 & \bf 58.3 & 59.3 & 60.5 & 72.9 & 35.6 & 55.14 \\
 & \textbf{SW-c1024}          & \bf 63.1 & 36.0 & 58.2 & \bf 59.7 & 60.9 & \bf 73.5 & 36.0 & 55.34 \\
 & \textbf{SW-1M}             & 62.8 & \bf 36.7 & 58.2 & 59.6 & 61.1 & 73.2 & 36.6 & \bf 55.45 \\
\hline
\end{tabular}
\end{center}
\end{table}

\begin{table}[H]
\caption{Zero-shot accuracy (higher better) of Llama-3.2-1B-Instruct with a \textbf{4-bit GuidedQuant body} and a tied head quantized at 2, 3 and 4 bits.
Details as in Table~\ref{tab:zeroshot-llama1b-instruct-b2}.
Best value per head group in bold.}
\label{tab:zeroshot-llama1b-instruct-b4}
\begin{center}
\scriptsize
\renewcommand{\arraystretch}{0.9}
\setlength{\tabcolsep}{3pt}
\begin{tabular}{llrrrrrrrr}
\multicolumn{1}{c}{\rule{0pt}{2.6ex}\bf Head} & \multicolumn{1}{c}{\bf Method} & \multicolumn{1}{c}{\bf ARC-e} & \multicolumn{1}{c}{\bf ARC-c} & \multicolumn{1}{c}{\bf HellaS} & \multicolumn{1}{c}{\bf LAMB} & \multicolumn{1}{c}{\bf Wino} & \multicolumn{1}{c}{\bf PIQA} & \multicolumn{1}{c}{\bf OBQA} & \multicolumn{1}{c}{\bf Mean}
\\ \hline
\rule{0pt}{2.6ex}16
 & fp16 body                  & 63.8 & 37.8 & 61.7 & 60.1 & 61.6 & 74.9 & 37.2 & 56.72 \\
16
 & fp16 head                  & 62.7 & 37.7 & 60.3 & 60.1 & 62.0 & 73.6 & 37.8 & 56.30 \\
\hline
\rule{0pt}{2.6ex}2
 & WS-c128                    & 60.5 & 36.6 & 58.9 & 53.6 & 61.5 & 72.2 & 35.8 & 54.16 \\
 & WS-c256                    & 61.0 & 37.0 & 59.0 & 53.9 & 60.7 & 72.1 & 36.2 & 54.27 \\
 & WS-c1024                   & 60.3 & 37.0 & 59.2 & 54.0 & 60.9 & 72.1 & \bf 37.2 & 54.39 \\
 & WS-1M                      & 59.5 & 35.5 & 59.5 & 55.0 & 60.1 & 72.5 & 36.8 & 54.12 \\
 & \textbf{SW-c128}           & 62.2 & 37.4 & 59.6 & 59.4 & 61.8 & 73.2 & 36.4 & 55.72 \\
 & \textbf{SW-c256}           & \bf 63.2 & \bf 38.7 & 59.5 & 59.2 & 61.3 & 73.3 & 35.4 & 55.81 \\
 & \textbf{SW-c1024}          & 61.2 & 37.5 & 59.3 & 59.2 & 61.6 & 72.7 & 35.0 & 55.23 \\
 & \textbf{SW-1M}             & 62.5 & 37.8 & \bf 59.7 & \bf 59.9 & \bf 62.0 & \bf 73.6 & 36.2 & \bf 55.96 \\
\cline{1-10}
\rule{0pt}{2.6ex}3
 & WS-c128                    & 62.2 & 38.5 & \bf 60.4 & 58.7 & 60.9 & \bf 73.8 & 37.4 & 55.98 \\
 & WS-c256                    & 62.7 & \bf 38.6 & 60.2 & 58.6 & 60.3 & 73.4 & 36.4 & 55.74 \\
 & WS-c1024                   & 62.5 & 38.2 & 60.2 & 58.2 & 60.5 & 73.1 & 36.4 & 55.60 \\
 & WS-1M                      & 62.7 & 37.9 & 60.3 & 58.4 & 60.5 & \bf 73.8 & 37.0 & 55.79 \\
 & \textbf{SW-c128}           & 62.2 & 37.9 & 60.3 & 59.5 & \bf 62.0 & 73.5 & 36.2 & 55.94 \\
 & \textbf{SW-c256}           & 62.5 & 36.9 & 60.2 & 60.0 & 61.8 & 73.4 & 36.6 & 55.93 \\
 & \textbf{SW-c1024}          & \bf 62.8 & 37.5 & 60.2 & \bf 60.2 & 61.1 & 73.3 & \bf 37.6 & 56.10 \\
 & \textbf{SW-1M}             & \bf 62.8 & 38.1 & 60.1 & 60.1 & \bf 62.0 & 73.6 & 36.6 & \bf 56.18 \\
\cline{1-10}
\rule{0pt}{2.6ex}4
 & WS-c128                    & \bf 63.2 & 37.0 & 60.3 & 59.5 & 62.0 & 73.7 & 37.8 & 56.23 \\
 & WS-c256                    & 62.6 & 37.5 & \bf 60.4 & 59.4 & 62.0 & 73.6 & 38.0 & 56.22 \\
 & WS-c1024                   & 63.1 & 37.8 & 60.3 & 59.7 & 62.0 & \bf 74.2 & 37.8 & \bf 56.41 \\
 & WS-1M                      & \bf 63.2 & 37.7 & 60.3 & 59.5 & 61.5 & 73.6 & \bf 38.4 & 56.30 \\
 & \textbf{SW-c128}           & 62.5 & 37.8 & 60.2 & 59.8 & \bf 62.5 & 73.4 & 37.6 & 56.27 \\
 & \textbf{SW-c256}           & 62.5 & \bf 38.0 & 60.2 & 59.9 & 62.4 & 74.0 & 37.4 & 56.35 \\
 & \textbf{SW-c1024}          & 62.9 & 37.5 & 60.3 & \bf 60.3 & 62.2 & 73.6 & 36.8 & 56.22 \\
 & \textbf{SW-1M}             & 62.4 & 37.8 & 60.3 & 60.2 & 61.9 & 73.7 & 37.8 & 56.29 \\
\hline
\end{tabular}
\end{center}
\end{table}

\subsection{Class-side statistic ablation}

\begin{table}[H]
\caption{Class-side statistic ablation, head-only isolation: the body is kept in fp16 and only the head matrix is quantized; tied models (Llama-3.2-1B, Qwen3-1.7B) retain an fp16 copy of the embedding.
All arms share the head, the damped $\Sigma_{X}$ Cholesky, the rate-matching routine and the $\epsilon = 0.1$ smoothing prior, so they differ only in the class-side statistic:
$\bar{p} = \E_{X}[\tilde{p}_{k}]$ is the model's average marginal;
$\pi$ is the empirical unigram frequency of token $k$ over the calibration window;
SW uses $\tilde{\lambda}_{k} = \E_{X}[\tilde{p}_{k}(1-\tilde{p}_{k})]$.
KL is measured against the all-fp16 model on identical token streams.
Every arm lands within $0.001$ bits of its nominal head rate.
Parentheses after each model name give its vocabulary size $V$; on PPL cells, the increase over the fp16 row of the same model.
Best value per column per head group in bold; ties bold both.}
\label{tab:class-side-statistics}
\begin{center}
\scriptsize
\renewcommand{\arraystretch}{0.85}
\setlength{\tabcolsep}{3pt}
\begin{tabular}{lrllll}
\multicolumn{1}{c}{\bf Statistic} & \multicolumn{1}{c}{\bf Head} & \multicolumn{1}{c}{\bf WT2 PPL ($\downarrow$)} & \multicolumn{1}{c}{\bf WT2 KL ($\downarrow$)} & \multicolumn{1}{c}{\bf C4 PPL ($\downarrow$)} & \multicolumn{1}{c}{\bf C4 KL ($\downarrow$)}
\\ \hline
\multicolumn{6}{c}{\rule{0pt}{2.6ex}\bf Llama-3.2-1B (128{,}256)} \\ \hline
\rule{0pt}{2.6ex}fp16 & 16 & 11.313 & 0 & 15.614 & 0 \\
\cline{1-6}
\rule{0pt}{2.6ex}$\bar{p}$ & 2 & \textbf{11.418 ($+$0.9\%)} & \textbf{0.00952} & 16.102 ($+$3.1\%) & 0.02959 \\
$\pi$ & 2 & 11.524 ($+$1.9\%) & 0.02146 & 16.549 ($+$6.0\%) & 0.05830 \\
\textbf{SW (ours)} & 2 & 11.419 ($+$0.9\%) & 0.00957 & \textbf{16.035 ($+$2.7\%)} & \textbf{0.02834} \\
\cline{1-6}
\rule{0pt}{2.6ex}$\bar{p}$ & 3 & 11.344 ($+$0.3\%) & 0.00213 & 15.719 ($+$0.7\%) & 0.00613 \\
$\pi$ & 3 & 11.356 ($+$0.4\%) & 0.00437 & 15.784 ($+$1.1\%) & 0.01184 \\
\textbf{SW (ours)} & 3 & \textbf{11.342 ($+$0.3\%)} & \textbf{0.00206} & \textbf{15.704 ($+$0.6\%)} & \textbf{0.00572} \\
\cline{1-6}
\rule{0pt}{2.6ex}$\bar{p}$ & 4 & \textbf{11.319 ($+$0.1\%)} & 0.00049 & \textbf{15.624 ($+$0.1\%)} & 0.00147 \\
$\pi$ & 4 & 11.327 ($+$0.1\%) & 0.00105 & 15.658 ($+$0.3\%) & 0.00280 \\
\textbf{SW (ours)} & 4 & 11.320 ($+$0.1\%) & \textbf{0.00047} & 15.640 ($+$0.2\%) & \textbf{0.00138} \\
\hline
\multicolumn{6}{c}{\rule{0pt}{3.2ex}\bf Qwen3-1.7B (151{,}936)} \\ \hline
\rule{0pt}{2.6ex}fp16 & 16 & 18.285 & 0 & 22.621 & 0 \\
\cline{1-6}
\rule{0pt}{2.6ex}$\bar{p}$ & 2 & \textbf{18.452 ($+$0.9\%)} & 0.00951 & 23.138 ($+$2.3\%) & 0.02110 \\
$\pi$ & 2 & 18.595 ($+$1.7\%) & 0.01912 & 23.494 ($+$3.9\%) & 0.04421 \\
\textbf{SW (ours)} & 2 & 18.460 ($+$1.0\%) & \textbf{0.00900} & \textbf{23.075 ($+$2.0\%)} & \textbf{0.01998} \\
\cline{1-6}
\rule{0pt}{2.6ex}$\bar{p}$ & 3 & \textbf{18.310 ($+$0.1\%)} & 0.00206 & 22.756 ($+$0.6\%) & 0.00468 \\
$\pi$ & 3 & 18.360 ($+$0.4\%) & 0.00389 & 22.823 ($+$0.9\%) & 0.00870 \\
\textbf{SW (ours)} & 3 & 18.317 ($+$0.2\%) & \textbf{0.00199} & \textbf{22.724 ($+$0.5\%)} & \textbf{0.00443} \\
\cline{1-6}
\rule{0pt}{2.6ex}$\bar{p}$ & 4 & 18.303 ($+$0.1\%) & 0.00046 & 22.636 ($+$0.1\%) & 0.00112 \\
$\pi$ & 4 & 18.304 ($+$0.1\%) & 0.00091 & 22.669 ($+$0.2\%) & 0.00203 \\
\textbf{SW (ours)} & 4 & \textbf{18.298 ($+$0.1\%)} & \textbf{0.00045} & \textbf{22.633 ($+$0.1\%)} & \textbf{0.00105} \\
\hline
\multicolumn{6}{c}{\rule{0pt}{3.2ex}\bf Llama-3.1-8B-Instruct (128{,}256)} \\ \hline
\rule{0pt}{2.6ex}fp16 & 16 & 8.388 & 0 & 14.189 & 0 \\
\cline{1-6}
\rule{0pt}{2.6ex}$\bar{p}$ & 2 & 8.441 ($+$0.6\%) & 0.00658 & \textbf{14.530 ($+$2.4\%)} & 0.02237 \\
$\pi$ & 2 & 8.501 ($+$1.3\%) & 0.01356 & 14.822 ($+$4.5\%) & 0.04139 \\
\textbf{SW (ours)} & 2 & \textbf{8.437 ($+$0.6\%)} & \textbf{0.00636} & 14.534 ($+$2.4\%) & \textbf{0.02140} \\
\cline{1-6}
\rule{0pt}{2.6ex}$\bar{p}$ & 3 & \textbf{8.398 ($+$0.1\%)} & 0.00140 & \textbf{14.252 ($+$0.4\%)} & 0.00465 \\
$\pi$ & 3 & 8.412 ($+$0.3\%) & 0.00274 & 14.293 ($+$0.7\%) & 0.00852 \\
\textbf{SW (ours)} & 3 & \textbf{8.398 ($+$0.1\%)} & \textbf{0.00135} & 14.258 ($+$0.5\%) & \textbf{0.00437} \\
\cline{1-6}
\rule{0pt}{2.6ex}$\bar{p}$ & 4 & \textbf{8.390 ($+$0.0\%)} & 0.00033 & 14.207 ($+$0.1\%) & 0.00113 \\
$\pi$ & 4 & 8.396 ($+$0.1\%) & 0.00066 & 14.225 ($+$0.3\%) & 0.00200 \\
\textbf{SW (ours)} & 4 & 8.391 ($+$0.0\%) & \textbf{0.00032} & \textbf{14.205 ($+$0.1\%)} & \textbf{0.00103} \\
\hline
\multicolumn{6}{c}{\rule{0pt}{3.2ex}\bf Qwen3-8B (151{,}936)} \\ \hline
\rule{0pt}{2.6ex}fp16 & 16 & 10.832 & 0 & 15.486 & 0 \\
\cline{1-6}
\rule{0pt}{2.6ex}$\bar{p}$ & 2 & 10.915 ($+$0.8\%) & \textbf{0.00643} & 15.699 ($+$1.4\%) & 0.01484 \\
$\pi$ & 2 & 10.965 ($+$1.2\%) & 0.01183 & 15.899 ($+$2.7\%) & 0.02919 \\
\textbf{SW (ours)} & 2 & \textbf{10.898 ($+$0.6\%)} & 0.00654 & \textbf{15.689 ($+$1.3\%)} & \textbf{0.01385} \\
\cline{1-6}
\rule{0pt}{2.6ex}$\bar{p}$ & 3 & \textbf{10.845 ($+$0.1\%)} & \textbf{0.00129} & 15.544 ($+$0.4\%) & 0.00328 \\
$\pi$ & 3 & 10.853 ($+$0.2\%) & 0.00251 & 15.577 ($+$0.6\%) & 0.00585 \\
\textbf{SW (ours)} & 3 & 10.848 ($+$0.1\%) & 0.00132 & \textbf{15.523 ($+$0.2\%)} & \textbf{0.00295} \\
\cline{1-6}
\rule{0pt}{2.6ex}$\bar{p}$ & 4 & \textbf{10.835 ($+$0.0\%)} & 0.00032 & \textbf{15.495 ($+$0.1\%)} & 0.00075 \\
$\pi$ & 4 & 10.839 ($+$0.1\%) & 0.00059 & 15.504 ($+$0.1\%) & 0.00132 \\
\textbf{SW (ours)} & 4 & 10.836 ($+$0.0\%) & \textbf{0.00031} & 15.501 ($+$0.1\%) & \textbf{0.00073} \\
\hline
\multicolumn{6}{c}{\rule{0pt}{3.2ex}\bf Qwen3-32B (151{,}936)} \\ \hline
\rule{0pt}{2.6ex}fp16 & 16 & 8.452 & 0 & 12.269 & 0 \\
\cline{1-6}
\rule{0pt}{2.6ex}$\bar{p}$ & 2 & 8.516 ($+$0.8\%) & 0.00709 & \textbf{12.475 ($+$1.7\%)} & 0.01663 \\
$\pi$ & 2 & 8.571 ($+$1.4\%) & 0.01461 & 12.703 ($+$3.5\%) & 0.03348 \\
\textbf{SW (ours)} & 2 & \textbf{8.498 ($+$0.5\%)} & \textbf{0.00703} & 12.479 ($+$1.7\%) & \textbf{0.01596} \\
\cline{1-6}
\rule{0pt}{2.6ex}$\bar{p}$ & 3 & 8.468 ($+$0.2\%) & 0.00153 & \textbf{12.303 ($+$0.3\%)} & 0.00347 \\
$\pi$ & 3 & 8.476 ($+$0.3\%) & 0.00297 & 12.356 ($+$0.7\%) & 0.00687 \\
\textbf{SW (ours)} & 3 & \textbf{8.464 ($+$0.1\%)} & \textbf{0.00147} & 12.315 ($+$0.4\%) & \textbf{0.00341} \\
\cline{1-6}
\rule{0pt}{2.6ex}$\bar{p}$ & 4 & \textbf{8.457 ($+$0.1\%)} & 0.00036 & 12.276 ($+$0.1\%) & 0.00087 \\
$\pi$ & 4 & 8.461 ($+$0.1\%) & 0.00072 & 12.289 ($+$0.2\%) & 0.00167 \\
\textbf{SW (ours)} & 4 & \textbf{8.457 ($+$0.1\%)} & \textbf{0.00035} & \textbf{12.275 ($+$0.0\%)} & \textbf{0.00081} \\
\hline
\end{tabular}
\end{center}
\end{table}

\clearpage

\section{Hessian sketching}
\label{app:hessian-sketching}

This appendix details the power iteration scheme used in \S\ref{sec:experiments} to compute optimized Kronecker factors of the head KL Hessian, adapted from the sketching subroutines of YAQA \citep{tseng_model-preserving_2026}.

\paragraph{Setup.}
We seek the best diagonal-Kronecker approximation
\begin{align}
    a^{\star}, B^{\star}
    = \argmin_{a \,\mathrm{diagonal}, \; B}
    \left\Vert
    \E_{X}\!\left[\tilde{\lambda}(X) \otimes XX^{\top}\right]
    - a \otimes B
    \right\Vert_{F}^{2},
    \label{eq:sketch-objective}
\end{align}
where $\tilde{\lambda}(X) = \diag(\tilde{p}\odot(1-\tilde{p}))$ is the softmax curvature under the smoothed output distribution $\tilde{p}$ of \eqref{eq:uniform-smoothing} with $\epsilon = 0.1$.
As noted by \citet{tseng_model-preserving_2026}, following \citet{van_loan_ubiquitous_2000}, the best Kronecker approximation is a rank-one approximation of a rearrangement of the target matrix, so alternating least squares (ALS) on the two factors is power iteration on that rearrangement and converges to the leading singular pair, which is unique up to the scale ambiguity $(ca, B/c)$ whenever the leading singular value is simple.
Since all our comparisons (cosine similarity to the fixed point, and the quantization lattice itself, through the coupling of $c_{\text{SW}}$) are invariant to this rescaling, we do not normalize the factors during the iteration.

\paragraph{Adapted updates.}
YAQA's Sketch~A estimates the per-token Hessian with a Monte-Carlo label sampled from the model's predictive distribution and a modified backward pass.
For the head layer both approximations are unnecessary: the layer's KL Hessian at a hidden state $X$ is available in closed form as $(\diag(\tilde{p}) - \tilde{p}\tilde{p}^{\top}) \otimes XX^{\top}$, requiring only a forward pass, and the token-independence bias of Sketch~A does not arise because no computation mixes tokens after the head.
We therefore replace the sampled label with the exact per-token curvature, which preserves the expectation of their estimator while removing its sampling variance.
Moreover, since $a$ is constrained to be diagonal, the off-diagonal coupling $-\tilde{p}\tilde{p}^{\top}$ is Frobenius-orthogonal to the ansatz $a \otimes B$, and the diagonal of the exact class-side Hessian is precisely $\tilde{\lambda} = \tilde{p}\odot(1-\tilde{p})$; hence each update below is the exact ALS step for \eqref{eq:sketch-objective}, not an approximation of it.
Writing $s(X) = \tilde{p}\odot(1-\tilde{p}) \in \sR^{K}$, one round consists of a class-side pass followed by a feature-side pass over the calibration set, mirroring the alternation order of the official YAQA release:
\begin{align}
    a &\leftarrow \frac{\E_{X}\!\left[ (X^{\top} B X)\, s(X) \right]}{\Vert B \Vert_{F}^{2}},
    &
    B &\leftarrow \frac{\E_{X}\!\left[ \langle a, s(X)\rangle\, XX^{\top} \right]}{\Vert a \Vert_{2}^{2}},
    \label{eq:sketch-updates}
\end{align}
where each update uses the other factor's most recent value.
Each pass costs one matrix-vector product with $B$ (or one rank-one accumulation into $B$) plus the head forward pass per token, and requires no backpropagation, in contrast to the general-layer setting of \citet{tseng_model-preserving_2026}.

\paragraph{Protocol.}
We cache the calibration hidden states once (WT2-train) and run all initializations of Table~\ref{tab:sw-fixed-point} on the identical cached states, so that differences reflect the initialization alone.
We run 8 rounds of \eqref{eq:sketch-updates}; the fixed point $(a^{\star}, B^{\star})$ is taken as the final iterate, and convergence is monitored by the cosine similarity of consecutive iterates, which exceeds $0.999999$ for every initialization from round 3 onward.
Note that the SoftWater factors $(\tilde{\lambda}, \Sigma_{X})$ are not merely a good initialization: round 0 of Table~\ref{tab:sw-fixed-point} shows they already agree with the fixed point to cosine similarity $0.976$--$0.992$, quantifying how close the analytical separable factorization is to the Frobenius-optimal one.

\end{document}